\documentclass{article}

\PassOptionsToPackage{numbers, compress}{natbib}

\newif\ifarxiv
\arxivtrue

\ifarxiv    \usepackage[main, final]{neurips_2026} 
\else   \usepackage[main, final]{neurips_2026} 
\fi

\usepackage[utf8]{inputenc}
\usepackage[T1]{fontenc}
\usepackage[pagebackref=true]{hyperref}
\usepackage{url}
\usepackage{booktabs}
\usepackage{amsfonts}
\usepackage{nicefrac}
\usepackage{microtype}
\usepackage{color,xcolor}
\usepackage{colortbl}
\usepackage{graphicx}
\usepackage{tikz}
\usepackage{amsmath,amssymb}
\usepackage{mathtools}
\usepackage[normalem]{ulem}
\usepackage{tabu}
\usepackage{multirow}
\usepackage{pgfplots,pgfplotstable}
\usepgfplotslibrary{fillbetween}
\usepackage{calc}
\usepackage{pbox}
\usepackage{algorithm}
\usepackage[noend]{algpseudocode}
\usepackage{caption}
\usepackage{afterpage}
\usepackage{subcaption}
\usepackage{wrapfig}
\usepackage{listings}
\usepackage{enumitem}
\usepackage[capitalise]{cleveref}
\usepackage{fancyvrb}
\usepackage{fvextra}

\definecolor{darkgreen}{rgb}{0.15, 0.75, 0.15}
\definecolor{mitblue}{rgb}{0.88,0.95,0.96}
\definecolor{lightblue}{RGB}{171, 219, 227}
\definecolor{citecolor}{rgb}{34,139,34}
\definecolor{mydarkblue}{RGB}{30,129,176}
\definecolor{mydarkgreen}{rgb}{0.12,0.7,0.12}
\definecolor{mydarkred}{rgb}{0.8,0.02,0.02}
\definecolor{mydarkorange}{rgb}{0.40,0.2,0.02}
\definecolor{mypurple}{RGB}{111,0,255}
\definecolor{myred}{RGB}{255,237,237}
\definecolor{mygold}{RGB}{238, 238, 228}
\definecolor{mydarkgray}{rgb}{0.66,0.66,0.66}
\definecolor{neutralgray}{RGB}{235, 237, 240}

\hypersetup{
    colorlinks=true,
    linkcolor=mydarkblue,
    citecolor=mydarkblue,
    filecolor=mydarkgreen,
    urlcolor=mydarkgreen,
}

\def\algname{HyperVAttention{}}
\def\algnameshort{HVA{}}

\pgfplotsset{compat=1.14}
\pgfplotsset{compat/show suggested version=false}
\setlist[itemize]{leftmargin=15pt}

\graphicspath{{./figures/}}
\newcommand{\figTeaser}{%
\begin{figure}[h]
    \vspace{-20pt}
    \centering
    \includegraphics[width=1\linewidth]{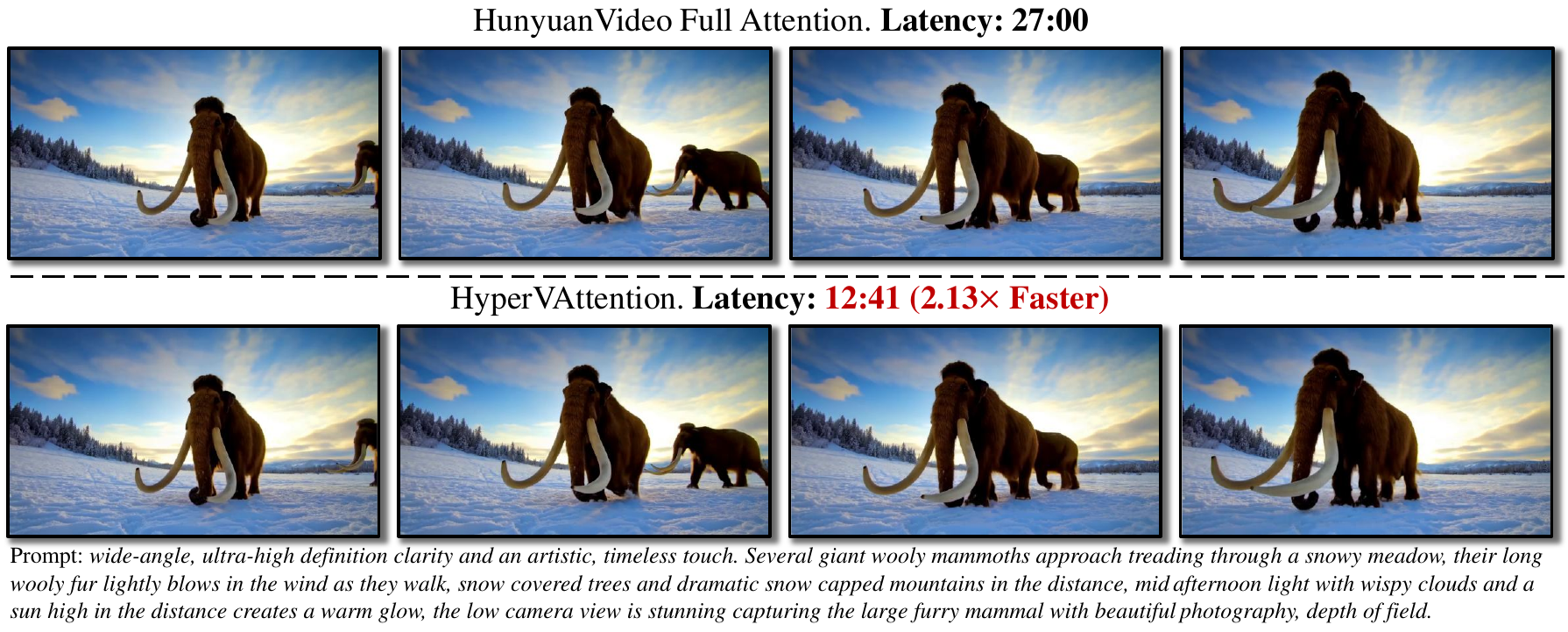}
    \vspace{-15pt}
    \caption{\algname~is a training-free sparse attention framework for video diffusion that achieves $2.13\times$ speedup on HunyuanVideo while preserving the visual fidelity of full attention.}
    \label{fig:teaser}
    \vspace{-10pt}
\end{figure}
}%

\newcommand{\figMotivationCombined}{%
\begin{figure*}[t]
    \centering
    \begin{minipage}[t]{0.73\linewidth}
        \vspace{0pt}
        \centering
        \subcaptionbox{\label{fig:motivation-cluster} Clustering Results of Video Latent Tokens.}{%
            \includegraphics[width=\linewidth]{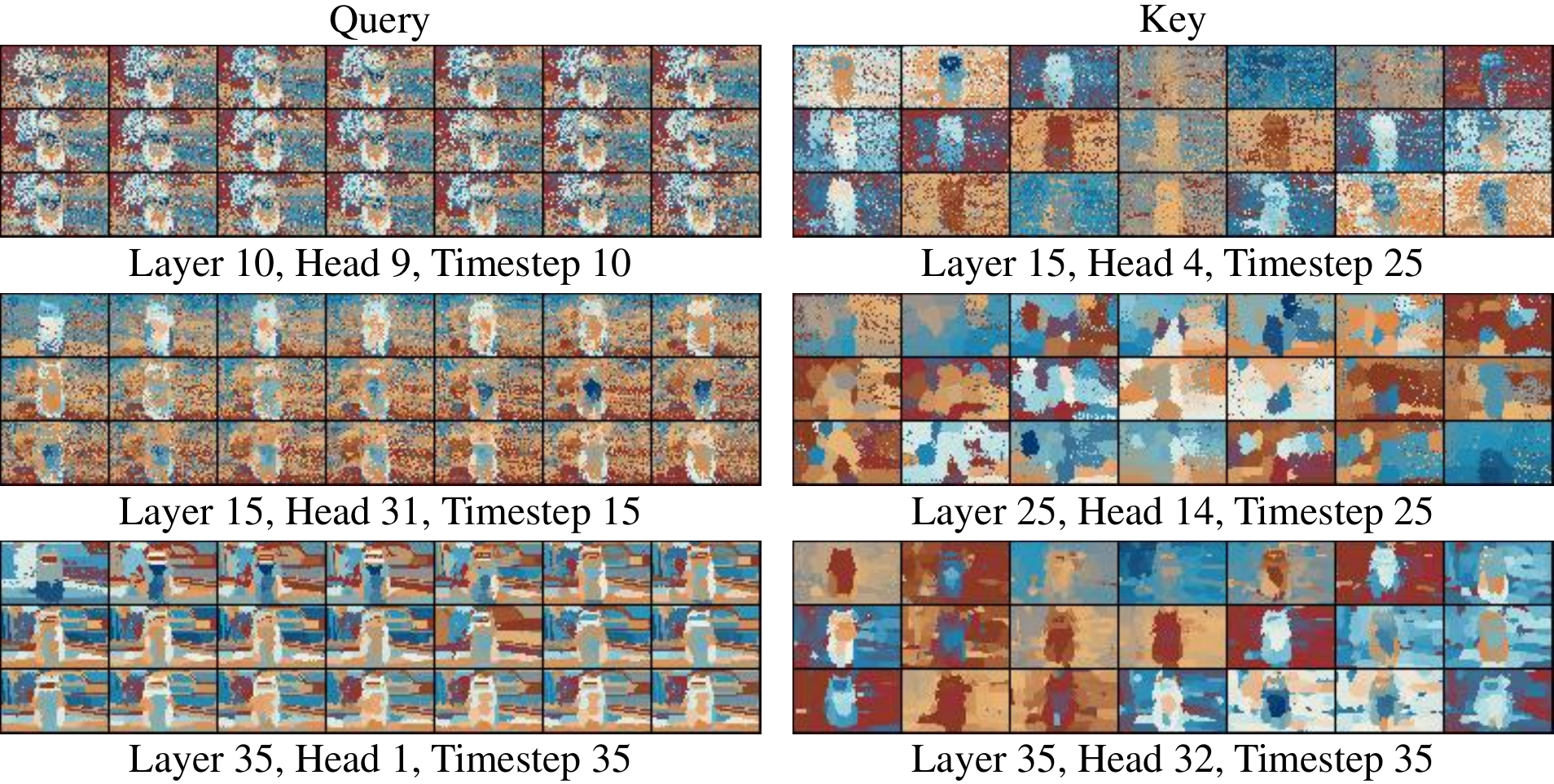}%
        }
    \end{minipage}
    \begin{minipage}[t]{0.0005\linewidth}
        \vspace{0pt}
        \centering
        \begin{tikzpicture}
            \draw[dashed, line width=0.9pt] (0,0) -- (0,-4.8);
        \end{tikzpicture}
    \end{minipage}
    \begin{minipage}[t]{0.24\linewidth}
        \vspace{0pt}
        \centering
        \subcaptionbox{\label{fig:motivation-change}Change Ratio}{%
            \includegraphics[width=\linewidth]{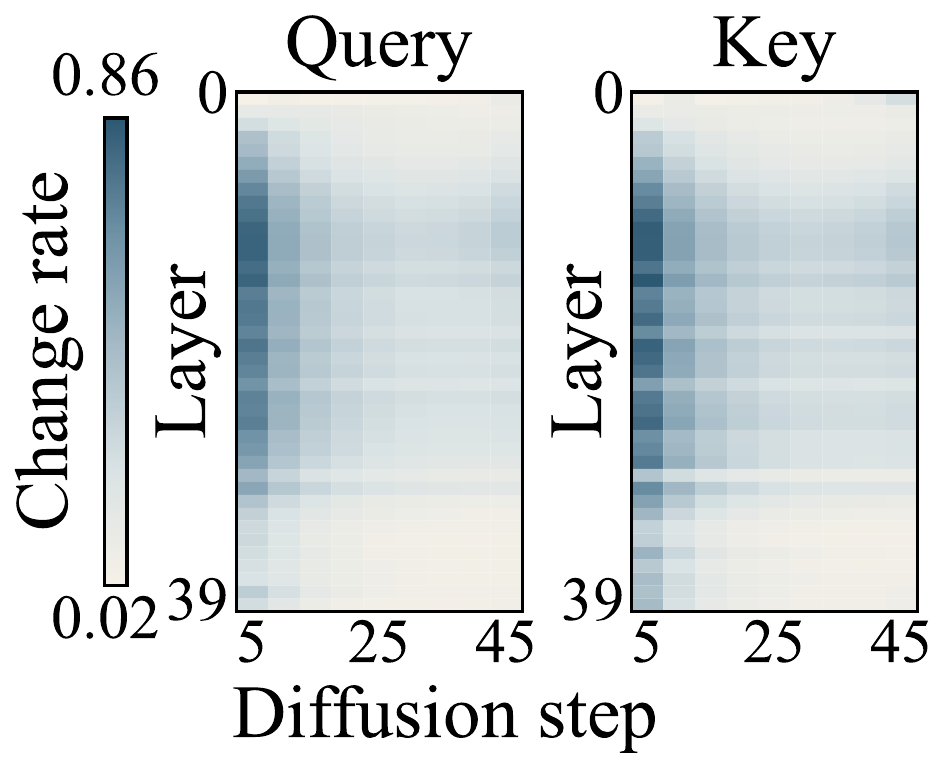}%
            \vspace{-3pt}
        }\\
        \subcaptionbox{\label{fig:motivation-lowe} Lowe's Ratio}{%
            \includegraphics[width=\linewidth]{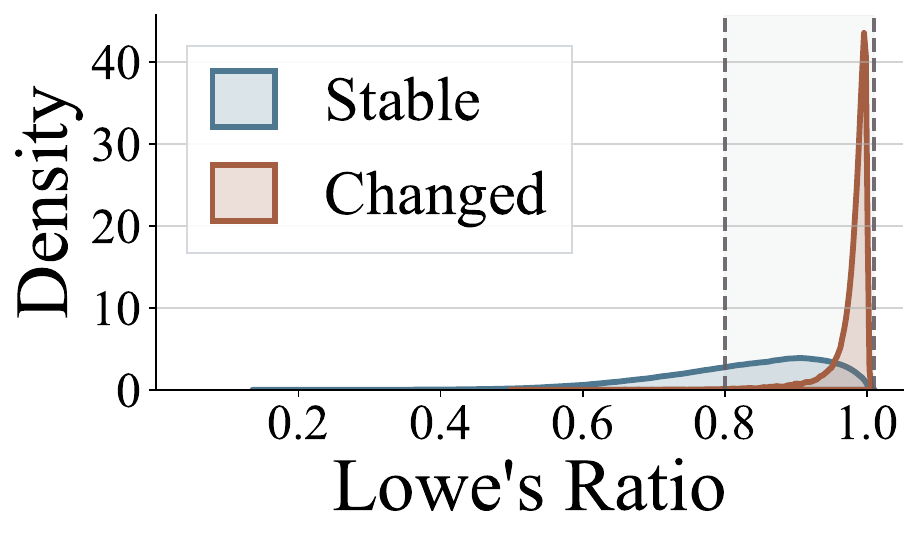}%
            \vspace{-2pt}
        }
    \end{minipage}
    \caption{(a) Clustering results of video latent tokens. Frames are ordered from the top left to the bottom right. (b) Ratio of cluster assignment changes over a five-step interval ($t$ vs. $t-5$). (c) Distributions of Lowe's ratio (see \cref{eq:lowes_ratio}) for reassigned and unchanged tokens.}
    \label{fig:motivation}
\end{figure*}
}%

\newcommand{\figLocalWindowClusteringKernelSpeed}{%
\begin{figure*}[t]
    \centering
    \makebox[\linewidth][l]{\hspace*{-0.005\linewidth}%
    \begin{minipage}[c]{0.5\linewidth}
        \centering
        \subcaptionbox{3D Local-Window Clustering\label{fig:lw_clustering_kernel_speed_lw_clustering}}{%
            \includegraphics[width=\linewidth]{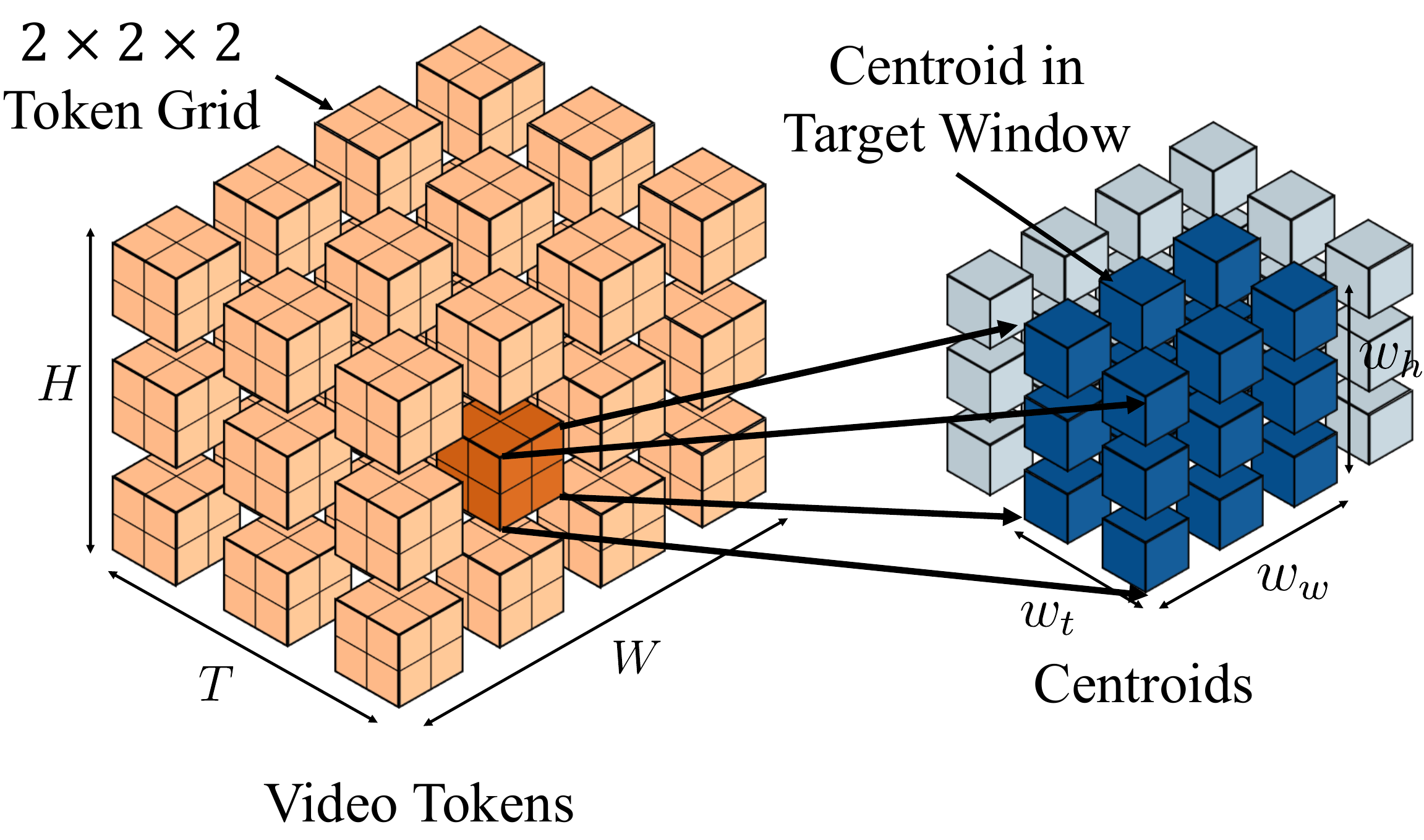}%
        }
    \end{minipage}
    \begin{minipage}[c]{0.1\linewidth}
        \centering
        \raisebox{0.4\columnwidth}{%
            \begin{tikzpicture}
                \draw[dashed, line width=0.8pt] (0,0.2) -- (0,-3.8);
            \end{tikzpicture}%
        }
    \end{minipage}
    \begin{minipage}[c]{0.31\linewidth}
        \centering
        \vspace{2pt}
        \subcaptionbox{Kernel Speed Comparison\label{fig:lw_clustering_kernel_speed_kernel}}{%
            \includegraphics[width=\linewidth]{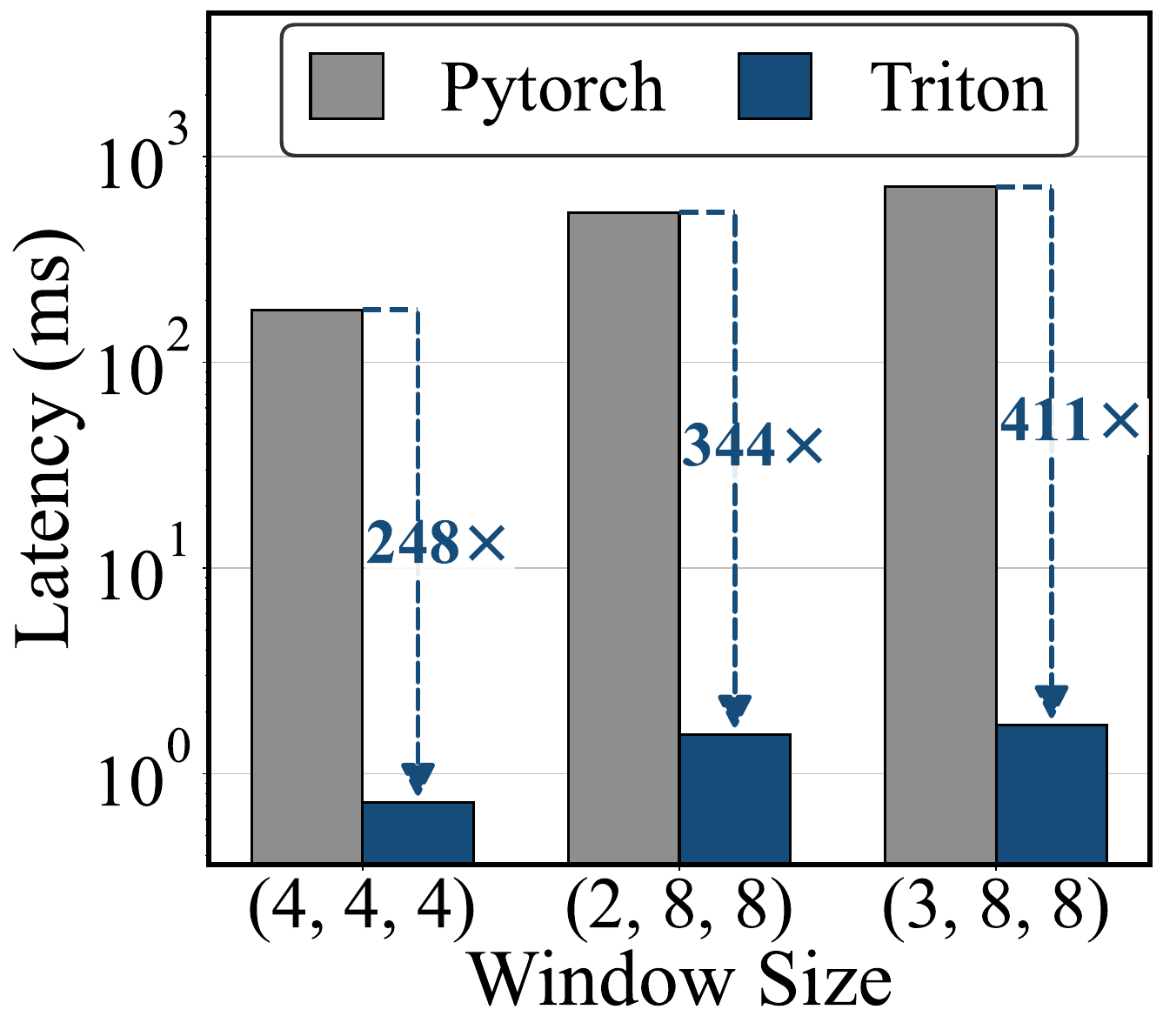}%
            \vspace{5pt}
        }
    \end{minipage}
    }%
    \caption{(a) Illustration of 3D Local-Window Clustering. (b) Latency comparison between  PyTorch and Triton implementations of 3D Local-Window Clustering varying window sizes.}
    \label{fig:lw_clustering_kernel_speed}
\end{figure*}
}%

\newcommand{\figClusterMerging}{%
\begin{figure}[t]
    \centering
    \vspace{-5pt}
    \includegraphics[width=1\linewidth]{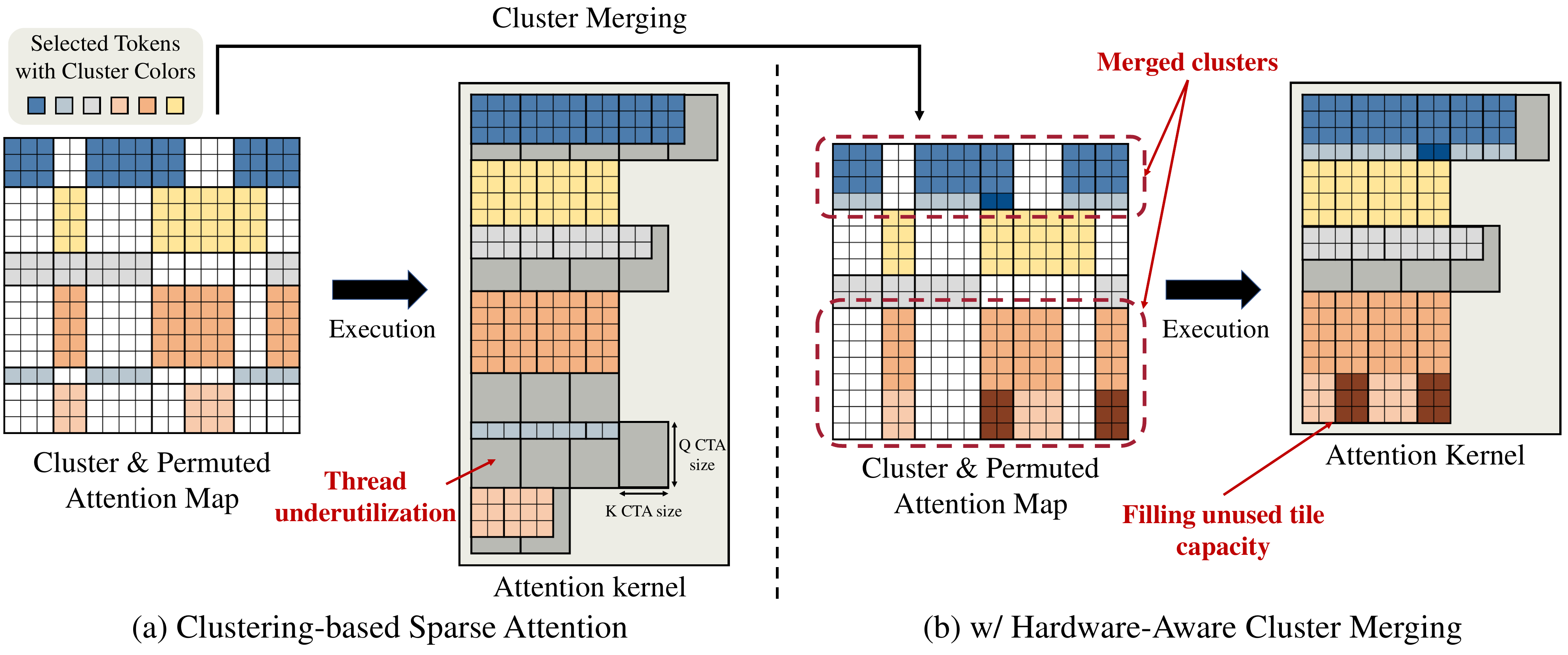}
    \caption{
        Illustration of how existing clustering-based sparse attention methods lead to thread underutilization. 
        (a) These methods form cluster-induced blocks, whose irregular sizes often misalign with fixed CTA tiles and leave padded regions idle. 
        (b) Merging similar clusters better matches cluster-induced blocks to CTA tiles, reducing tiled execution cost. Darker regions show additional tokens placed in unused tile capacity, improving approximation fidelity without increasing CTA tiles.
    }
    \label{fig:cluster_merging}
    \vspace{-5pt}
\end{figure}
}%

\newcommand{\figLatency}{%
\begin{figure}[t]
    \centering
    \vspace{-5pt}
    \captionsetup[subfigure]{font=small,skip=1pt,margin={0.08\linewidth,0pt}}
    \makebox[\linewidth][l]{\hspace*{-0.01\linewidth}%
    \subcaptionbox{End-to-End Latency Breakdown\label{fig:speed_latency_breakdown}}[0.354\linewidth]{%
        \begin{minipage}[t][0.292\columnwidth][t]{\linewidth}
            \centering
            \raisebox{0.065\columnwidth}{\includegraphics[width=\linewidth]{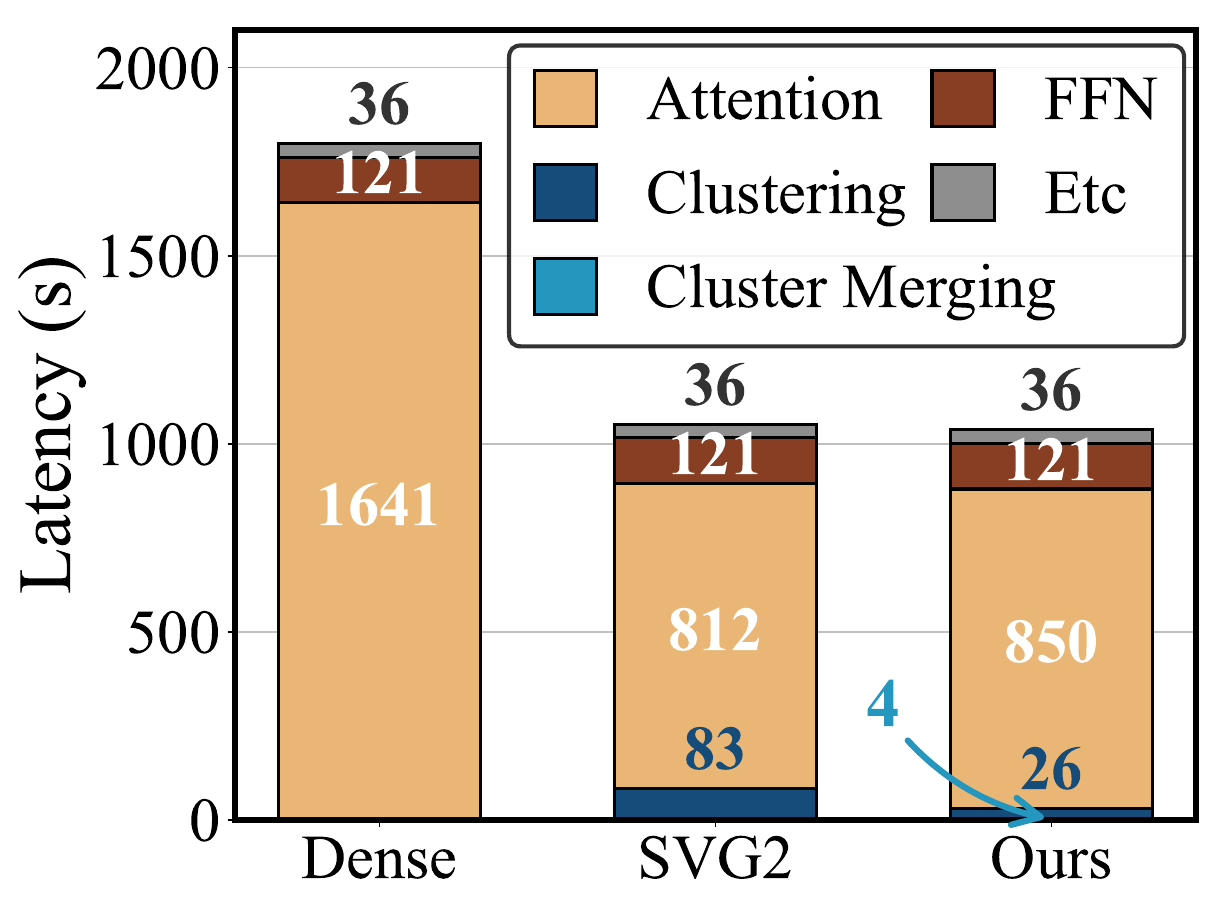}}
        \end{minipage}%
    }%
    \subcaptionbox{Clustering Efficiency\label{fig:speed_clustering_efficiency}}[0.325\linewidth]{%
        \begin{minipage}[t][0.292\columnwidth][t]{\linewidth}
            \centering
            \includegraphics[width=\linewidth]{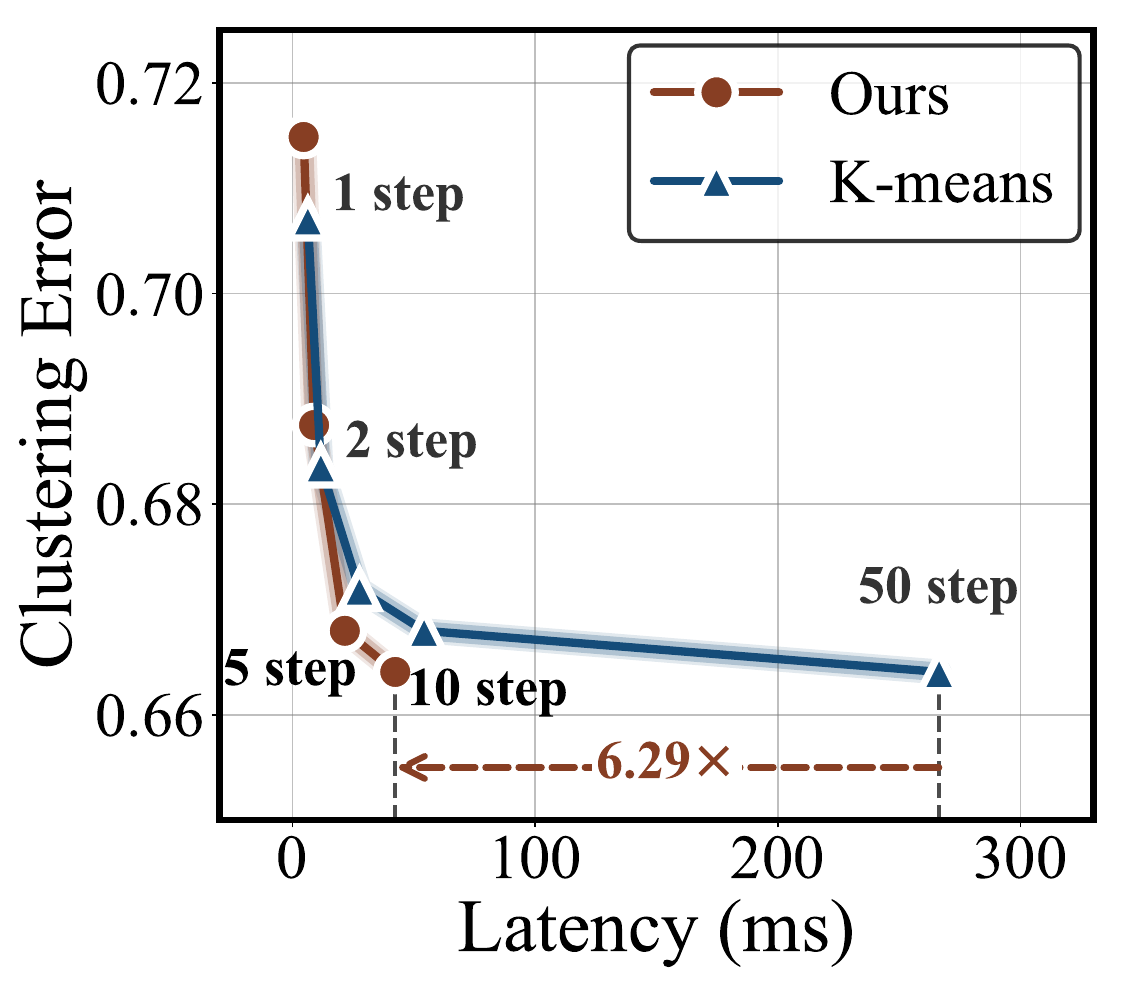}
        \end{minipage}%
    }%
    \subcaptionbox{Density Comparison\label{fig:speed_density_comparison}}[0.3115\linewidth]{%
        \begin{minipage}[t][0.292\columnwidth][t]{\linewidth}
            \centering
            \raisebox{0.01\columnwidth}{\includegraphics[width=\linewidth]{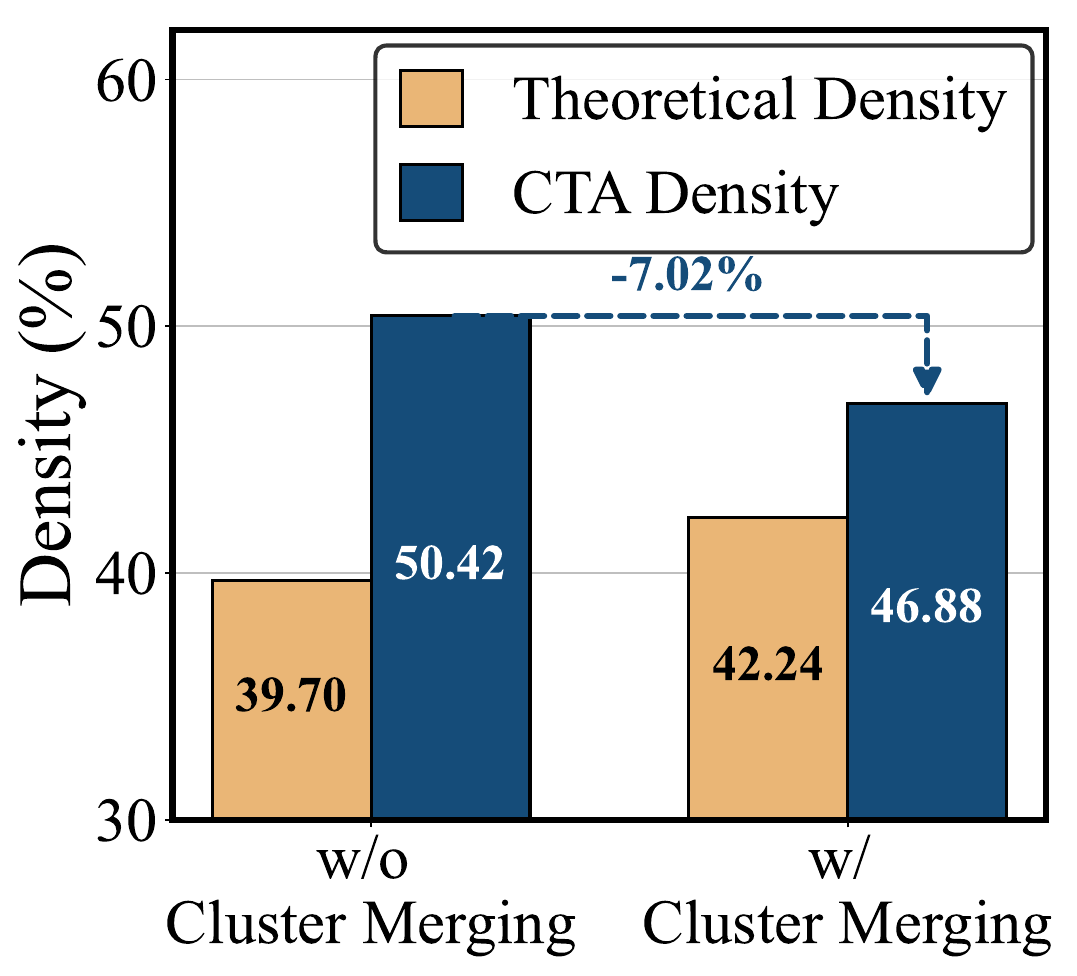}}
        \end{minipage}%
    }
    }%
    \caption{
        Efficiency analysis on diverse component of \algname.
        }
    \label{fig:speed_e2e}
    \vspace{-5pt}
\end{figure}
}%

\newcommand{\figSupWanItoV}{%
\begin{figure}[h]
    \centering
    \includegraphics[width=1\linewidth]{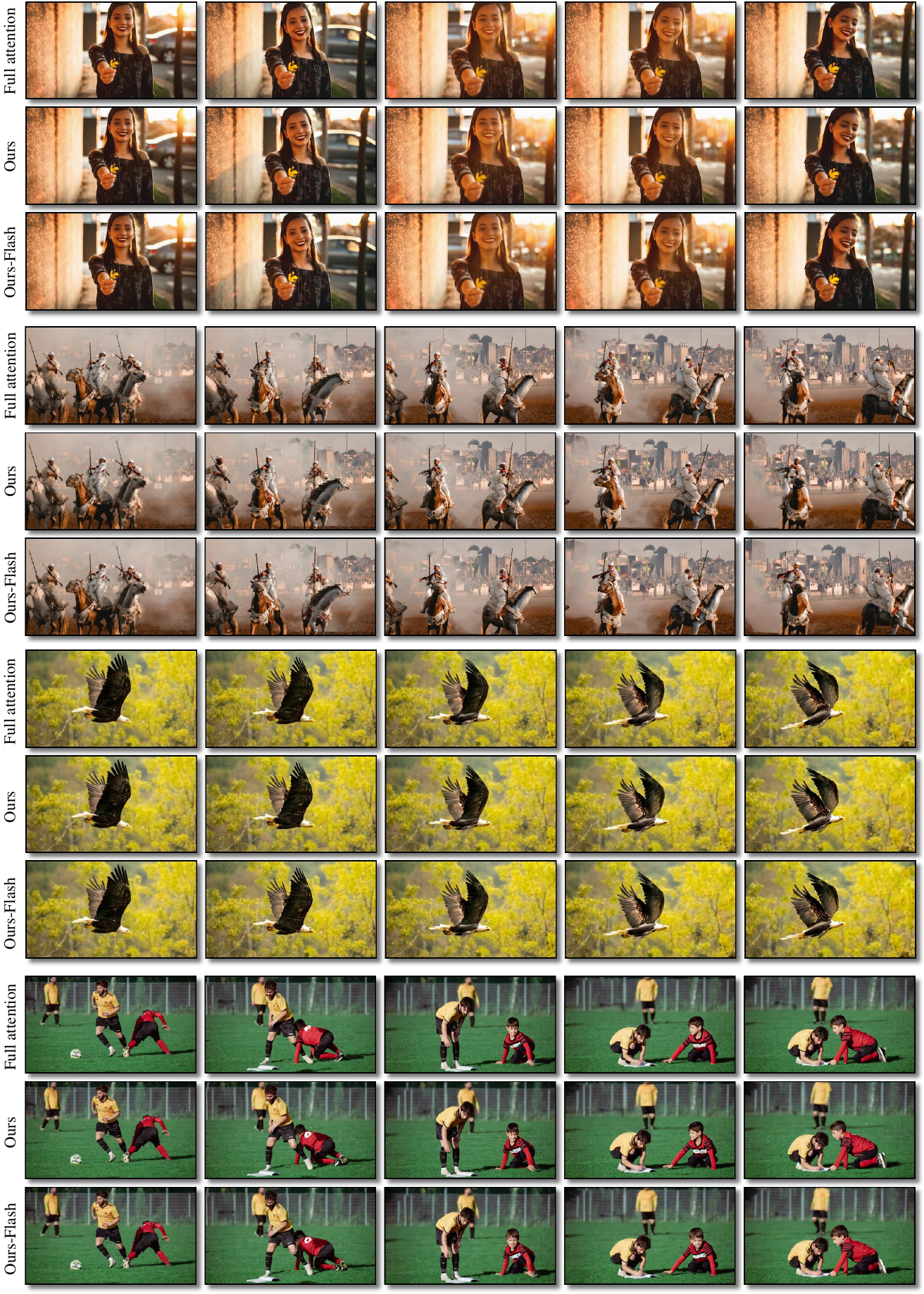}
    \caption{Visualization of generated videos from Wan2.2 on Image-to-Video generation.}
    \label{fig:fig:sup_wan_i_to_v}
\end{figure}
}%

\newcommand{\figSupWanTtoV}{%
\begin{figure}[h]
    \centering
    \includegraphics[width=1\linewidth]{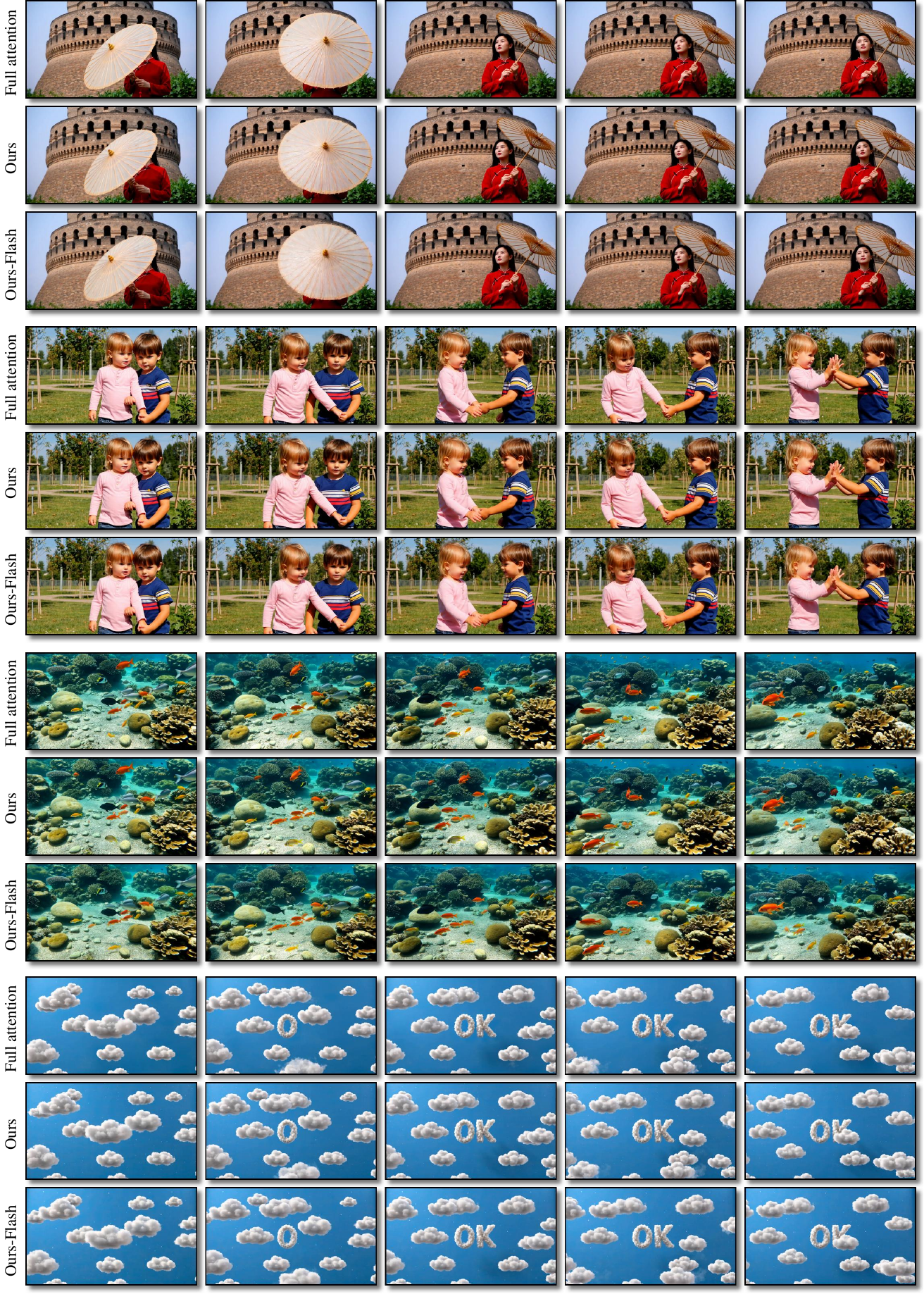}
    \caption{Visualization of generated videos from Wan2.2 on Text-to-Video generation.}
    \label{fig:fig:sup_wan_t_to_v}
\end{figure}
}%

\newcommand{\figSupHunyuanTtoV}{%
\begin{figure}[h]
    \centering
    \includegraphics[width=1\linewidth]{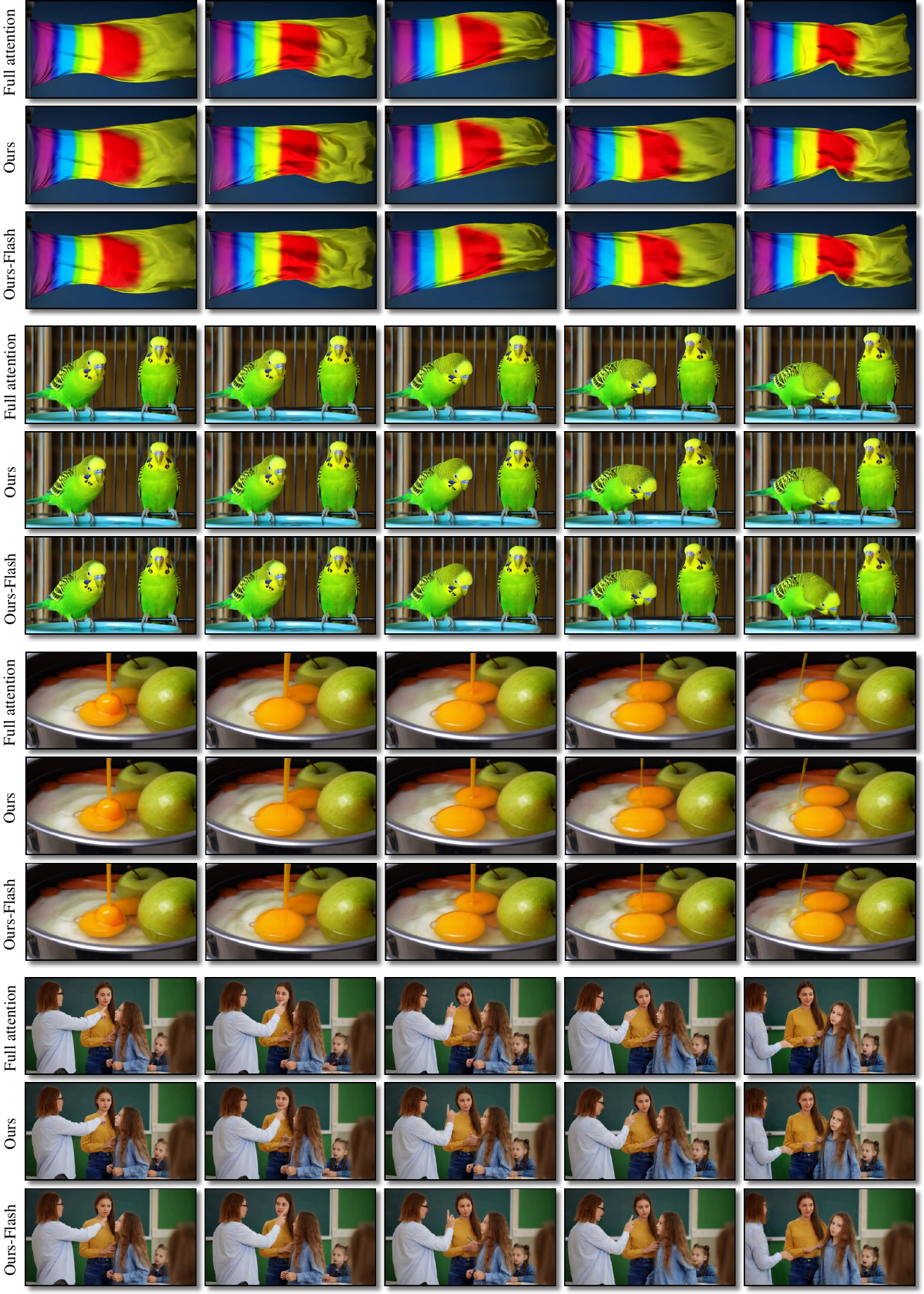}
    \caption{Visualization of generated videos from HunyuanVideo on Text-to-Video generation.}
    \label{fig:fig:sup_hunyuan_t_to_v}
\end{figure}
}%

\newcommand{\tabMain}{
\begin{table*}[t]
\centering
\vspace{-5pt}
\scriptsize
\caption{Quantitative comparison of quality and efficiency across various methods. \textbf{Bold} and \underline{underline} denote the best and second-best results, respectively.}
\label{tab:main}
\renewcommand{\arraystretch}{1.25}
\resizebox{1\textwidth}{!}{%
\begin{tabular}{lccccccc}
\toprule
\multicolumn{1}{c}{\multirow{2}{*}{\textbf{Methods}}} & \multicolumn{5}{c}{\textbf{Quality}} & \multicolumn{2}{c}{\textbf{Efficiency}} \\
\cmidrule(lr){2-6}\cmidrule(lr){7-8}
 & \textbf{PSNR} $\uparrow$ & \textbf{SSIM} $\uparrow$ & \textbf{LPIPS} $\downarrow$ & \textbf{VBench} $\uparrow$ & \textbf{LLM-as-a-Judge} $\uparrow$ & \textbf{Density} $\downarrow$ & \textbf{Speedup} $\uparrow$ \\
\midrule
\rowcolor{neutralgray}
\multicolumn{8}{c}{\textbf{Wan 2.2} (\textit{14B, Image-to-Video, 81 Frames, 720p, $n=$75{,}600})} \\
\addlinespace[2pt]
Full Attention & - & - & - & 0.816 & 8.24 & 100\% & $1\times$ \\
\addlinespace[-1pt]
\cmidrule(lr){1-8}
\addlinespace[-1pt]
SpargeAttn              & 24.95 & 0.869 & 0.106 & \underline{0.815} & \underline{8.11} & 30.15\% & 1.77$\times$ \\ 
SVG      & 23.06 & 0.817 & 0.143 & 0.814 & 7.47 & 34.29\% &  1.76$\times$ \\
SVG2      & 25.55 & 0.894 & 0.095 & 0.814 & 7.97 & 29.54\% & 1.80$\times$ \\
\cellcolor{mygold}\algnameshort      & \cellcolor{mygold}\textbf{26.84} & \cellcolor{mygold}\textbf{0.916} & \cellcolor{mygold}\textbf{0.077} & \cellcolor{mygold}\textbf{0.816} & \cellcolor{mygold}\textbf{8.15} & \cellcolor{mygold}36.95\% & \cellcolor{mygold}\underline{1.80$\times$} \\
\cellcolor{mygold}\algnameshort-Flash & \cellcolor{mygold}\underline{25.92} & \cellcolor{mygold}\underline{0.903} & \cellcolor{mygold}\underline{0.090} & \cellcolor{mygold}0.814 & \cellcolor{mygold}7.95 & \cellcolor{mygold}33.47\% & \cellcolor{mygold}\textbf{1.86$\times$} \\
\midrule
\rowcolor{neutralgray}
\multicolumn{8}{c}{\textbf{Wan 2.2} (\textit{14B, Text-to-Video, 81 Frames, 720p, $n=$75{,}600})} \\
\addlinespace[2pt]
Full Attention & - & - & - & 0.845 & 9.20 & 100\% & $1\times$ \\
\addlinespace[-1pt]
\cmidrule(lr){1-8}
\addlinespace[-1pt]
SpargeAttn              & 21.69 & 0.784 & 0.167 & 0.845 & 9.08 & 30.15\% & 1.77$\times$ \\
SVG      & 20.16 & 0.717 & 0.227 & \textbf{0.846} & 9.05 & 34.29\% & \underline{1.78$\times$} \\
SVG2      & 24.33 & 0.866 & 0.117 & 0.843 & 9.15 & 33.57\% & 1.71$\times$ \\
\cellcolor{mygold}\algnameshort      & \cellcolor{mygold}\textbf{25.50} & \cellcolor{mygold}\textbf{0.894} & \cellcolor{mygold}\textbf{0.094} & \cellcolor{mygold}\underline{0.845} & \cellcolor{mygold}\textbf{9.35} & \cellcolor{mygold}42.05\% & \cellcolor{mygold}1.72$\times$ \\
\cellcolor{mygold}\algnameshort-Flash & \cellcolor{mygold}\underline{24.42} & \cellcolor{mygold}\underline{0.867} & \cellcolor{mygold}\underline{0.116} & \cellcolor{mygold}0.843 & \cellcolor{mygold}\underline{9.18} & \cellcolor{mygold}36.72\% & \cellcolor{mygold}\textbf{1.81$\times$}  \\
\midrule

\rowcolor{neutralgray}
\multicolumn{8}{c}{\textbf{HunyuanVideo} (\textit{13B, Text-to-Video, 129 Frames, 720p, $n=$118{,}800})} \\
\addlinespace[2pt]
Full Attention & - & - & - & 0.821 & 9.05 & 100\% & $1\times$ \\
\addlinespace[-1pt]
\cmidrule(lr){1-8}
\addlinespace[-1pt]
SpargeAttn              & 28.12 & 0.894 & 0.103 & \textbf{0.822} & \underline{8.96} & 30.19\% & 1.94$\times$ \\
SVG      & 27.20 & 0.872 & 0.124 & 0.819 & 8.86 & 29.70\% & 1.91$\times$ \\
SVG2      & 32.66 & 0.947 & 0.056 & 0.817 & 8.89 & 26.27\% & 2.12$\times$ \\
\cellcolor{mygold}\algnameshort       & \cellcolor{mygold}\textbf{33.97} & \cellcolor{mygold}\textbf{0.958} & \cellcolor{mygold}\textbf{0.046} & \cellcolor{mygold}\underline{0.819} & \cellcolor{mygold}\textbf{9.18} & \cellcolor{mygold}31.34\% & \cellcolor{mygold}\underline{2.13$\times$} \\
\cellcolor{mygold}\algnameshort-Flash      & \cellcolor{mygold}\underline{32.92} & \cellcolor{mygold}\underline{0.949} & \cellcolor{mygold}\underline{0.055} & \cellcolor{mygold}0.818 & \cellcolor{mygold}8.92 & \cellcolor{mygold}28.35\% & \cellcolor{mygold}\textbf{2.23$\times$} \\

\bottomrule
\end{tabular}
}
\vspace{-5pt}
\end{table*}
}

\newcommand{\tabConfig}{
\begin{table*}[h]
\centering
\caption{Hyperparameters for all experiments including ours and baselines.}
\label{tab:hyperparams}
\renewcommand{\arraystretch}{1.3}
\resizebox{1\textwidth}{!}{%
\begin{tabular}{lccccccccccc}
\toprule
\textbf{Methods} & \textbf{Time warmup} & \textbf{Layer warmup} & \textbf{Q grid size} & \textbf{K grid size} & \textbf{Q window size} & \textbf{K window size} & $N_{q}$ & $N_{k}$ & \textbf{Top-\textit{p}} & \textbf{Min-\textit{kc}} & \textbf{Density} \\
\midrule

\rowcolor{neutralgray}
\multicolumn{12}{c}{\textbf{Wan 2.2} (\textit{14B, Image-to-Video, 81 Frames, 720p, $n=$75,600})} \\
\addlinespace[2pt]
SpargeAttn   & $10/50$ & $1/40$ & -- & -- & -- & -- & -- & -- & -- & -- & 0.3 \\ 
SVG          & $10/50$ & $1/40$ & -- & -- & -- & -- & -- & -- & -- & -- & 0.3 \\ 
SVG2         & $10/50$ & $1/40$ & -- & -- & -- & -- & $300$ & $1000$ & $0.9$ & $0.1$ & -- \\
\cellcolor{mygold}\algnameshort         & \cellcolor{mygold}$10/50$ & \cellcolor{mygold}$1/40$ & \cellcolor{mygold}$(7, 6, 6)$ & \cellcolor{mygold}$(6, 4, 4)$ & \cellcolor{mygold}$(3, 8, 8)$ & \cellcolor{mygold}$(3, 8, 8)$ & \cellcolor{mygold}$336$ & \cellcolor{mygold}$960$ & \cellcolor{mygold}$0.9$ & \cellcolor{mygold}$0.18$ & \cellcolor{mygold}-- \\
\cellcolor{mygold}\algnameshort-Flash   & \cellcolor{mygold}$10/50$ & \cellcolor{mygold}$1/40$ & \cellcolor{mygold}$(7, 6, 6)$ & \cellcolor{mygold}$(6, 4, 4)$ & \cellcolor{mygold}$(3, 8, 8)$ & \cellcolor{mygold}$(3, 8, 8)$ & \cellcolor{mygold}$336$ & \cellcolor{mygold}$960$ & \cellcolor{mygold}$0.9$ & \cellcolor{mygold}$0.12$ & \cellcolor{mygold}-- \\
\midrule

\rowcolor{neutralgray}
\multicolumn{12}{c}{\textbf{Wan 2.2} (\textit{14B, Text-to-Video, 81 Frames, 720p, $n=$75,600})} \\
\addlinespace[2pt]
SpargeAttn   & $10/50$ & $1/40$ & -- & -- & -- & -- & -- & -- & -- & -- & 0.3 \\ 
SVG          & $10/50$ & $1/40$ & -- & -- & -- & -- & -- & -- & -- & -- & 0.3 \\ 
SVG2         & $10/50$ & $1/40$ & -- & -- & -- & -- & $300$ & $1000$ & $0.9$ & $0.1$ & -- \\
\cellcolor{mygold}\algnameshort         & \cellcolor{mygold}$10/50$ & \cellcolor{mygold}$1/40$ & \cellcolor{mygold}$(7, 6, 6)$ & \cellcolor{mygold}$(6, 4, 4)$ & \cellcolor{mygold}$(3, 8, 8)$ & \cellcolor{mygold}$(3, 8, 8)$ & \cellcolor{mygold}$336$ & \cellcolor{mygold}$960$ & \cellcolor{mygold}$0.9$ & \cellcolor{mygold}$0.2$ & \cellcolor{mygold}-- \\
\cellcolor{mygold}\algnameshort-Flash   & \cellcolor{mygold}$10/50$ & \cellcolor{mygold}$1/40$ & \cellcolor{mygold}$(7, 6, 6)$ & \cellcolor{mygold}$(6, 4, 4)$ & \cellcolor{mygold}$(3, 8, 8)$ & \cellcolor{mygold}$(3, 8, 8)$ & \cellcolor{mygold}$336$ & \cellcolor{mygold}$960$ & \cellcolor{mygold}$0.9$ & \cellcolor{mygold}$0.1$ & \cellcolor{mygold}-- \\
\midrule

\rowcolor{neutralgray}
\multicolumn{12}{c}{\textbf{HunyuanVideo} (\textit{13B, Text-to-Video, 129 Frames, 720p, $n=$118,800})} \\
\addlinespace[2pt]
SpargeAttn   & $10/50$ & $1/20, 1/40$ & -- & -- & -- & -- & -- & -- & -- & -- & 0.3 \\ 
SVG          & $10/50$ & $1/20, 1/40$ & -- & -- & -- & -- & -- & -- & -- & -- & 0.3 \\ 
SVG2         & $10/50$ & $1/20, 1/40$ & -- & -- & -- & -- & $400$ & $1000$ & $0.9$ & $0.1$ & -- \\
\cellcolor{mygold}\algnameshort         & \cellcolor{mygold}$10/50$ & \cellcolor{mygold}$1/20, 1/40$ & \cellcolor{mygold}$(11, 5, 5)$ & \cellcolor{mygold}$(11, 3, 3)$ & \cellcolor{mygold}$(3, 8, 8)$ & \cellcolor{mygold}$(3, 8, 8)$ & \cellcolor{mygold}$432$ & \cellcolor{mygold}$1215$ & \cellcolor{mygold}$0.9$ & \cellcolor{mygold}$0.15$ & \cellcolor{mygold}-- \\
\cellcolor{mygold}\algnameshort-Flash   & \cellcolor{mygold}$10/50$ & \cellcolor{mygold}$1/20, 1/40$ & \cellcolor{mygold}$(11, 5, 5)$ & \cellcolor{mygold}$(11, 3, 3)$ & \cellcolor{mygold}$(3, 8, 8)$ & \cellcolor{mygold}$(3, 8, 8)$ & \cellcolor{mygold}$432$ & \cellcolor{mygold}$1215$ & \cellcolor{mygold}$0.9$ & \cellcolor{mygold}$0.1$ & \cellcolor{mygold}-- \\

\bottomrule
\end{tabular}
}
\end{table*}
}

\newcommand{\tabLLMJudge}{
\begin{table*}[h]
\centering
\caption{Quantitative comparison across different dimensions for video generation tasks. \textbf{Bold} and \underline{underline} denote the best and second-best results among sparse methods, respectively.}
\label{tab:video_results}
\renewcommand{\arraystretch}{1.4}
\resizebox{1\textwidth}{!}{%
\begin{tabular}{l|ccc|c}
\toprule
\textbf{Methods} & \textbf{Subject Cons.} $\uparrow$ & \textbf{Aesthetic Qual.} $\uparrow$ & \textbf{Temp. Flickering} $\uparrow$ & \textbf{Overall Score} $\uparrow$ \\
\midrule
\rowcolor{neutralgray}
\multicolumn{5}{c}{\textbf{Wan 2.2} (\textit{14B, Image-to-Video, 81 Frames, 720p, $n=$75,600})} \\
\addlinespace[2pt]
Full Attention & 7.49 & 8.68 & 8.55 & 8.24 \\
\addlinespace[-1pt]
\cmidrule(lr){1-5}
\addlinespace[-1pt]
SpargeAttn              & \textbf{7.56} & 8.13 & \textbf{8.64} & \underline{8.11} \\
SVG2                    & 7.40 & \underline{8.27} & 8.24 & 7.97 \\
\cellcolor{mygold}\algnameshort       & \cellcolor{mygold}\underline{7.42} & \cellcolor{mygold}\textbf{8.43} & \cellcolor{mygold}\underline{8.60} & \cellcolor{mygold}\textbf{8.15} \\
\cellcolor{mygold}\algnameshort-Flash & \cellcolor{mygold}7.35 & \cellcolor{mygold}8.20 & \cellcolor{mygold}8.30 & \cellcolor{mygold}7.95 \\
\midrule
\rowcolor{neutralgray}
\multicolumn{5}{c}{\textbf{Wan 2.2} (\textit{14B, Text-to-Video, 81 Frames, 720p, $n=$75,600})} \\
\addlinespace[2pt]
Full Attention & 9.29 & 8.89 & 9.42 & 9.20 \\
\addlinespace[-1pt]
\cmidrule(lr){1-5}
\addlinespace[-1pt]
SpargeAttn              & \underline{9.21} & 8.62 & 9.41 & 9.08 \\
SVG2                    & 9.19 & 8.84 & 9.42 & 9.15 \\
\cellcolor{mygold}\algnameshort & \cellcolor{mygold}\textbf{9.48} & \cellcolor{mygold}\textbf{8.95} & \cellcolor{mygold}\textbf{9.62} & \cellcolor{mygold}\textbf{9.35} \\
\cellcolor{mygold}\algnameshort-Flash & \cellcolor{mygold}9.20 & \cellcolor{mygold}\underline{8.89} & \cellcolor{mygold}\underline{9.45} & \cellcolor{mygold}\underline{9.18} \\
\midrule
\rowcolor{neutralgray}
\multicolumn{5}{c}{\textbf{HunyuanVideo} (\textit{13B, Text-to-Video, 129 Frames, 720p, $n=$118,800})} \\
\addlinespace[2pt]
Full Attention & 9.09 & 8.65 & 9.41 & 9.05 \\
\addlinespace[-1pt]
\cmidrule(lr){1-5}
\addlinespace[-1pt]
SpargeAttn              & 8.88 & \underline{8.72} & \underline{9.28} & \underline{8.96} \\
SVG2                    & 8.88 & 8.69 & 9.10 & 8.89 \\
\cellcolor{mygold}\algnameshort & \cellcolor{mygold}\textbf{9.15} & \cellcolor{mygold}\textbf{8.84} & \cellcolor{mygold}\textbf{9.55} & \cellcolor{mygold}\textbf{9.18} \\
\cellcolor{mygold}\algnameshort-Flash & \cellcolor{mygold}\underline{8.90} & \cellcolor{mygold}8.71 & \cellcolor{mygold}9.15 & \cellcolor{mygold}8.92 \\
\bottomrule
\end{tabular}
}
\end{table*}
}

\newcommand{\tabComponentAblation}{
\begin{table}[!htbp]
\centering
\small
\caption{Cumulative component ablation using HunyuanVideo on an H200 GPU.}
\label{tab:component_ablation}
\renewcommand{\arraystretch}{1.4}
\resizebox{0.9\textwidth}{!}{%
\begin{tabular}{lccccc}
\toprule
\textbf{Method} & \textbf{PSNR} $\uparrow$ & \textbf{SSIM} $\uparrow$ & \textbf{LPIPS} $\downarrow$ & \textbf{Density} $\downarrow$ & \textbf{Speedup} $\uparrow$ \\
\midrule
Local Window Clustering & 34.44 & 0.964 & 0.036 & 29.74\% & 2.05$\times$ \\
+ Hybrid Update & 34.59 & 0.965 & 0.035 & 28.83\% & 2.06$\times$ \\
\rowcolor{mygold}
+ Cluster Merging (\algnameshort) & \textbf{34.73} & \textbf{0.967} & \textbf{0.034} & 31.33\% & \textbf{2.11$\times$} \\
\bottomrule
\end{tabular}
}
\end{table}
}

\newcommand{\tabMergingAblation}{
\begin{table}[!htbp]
\centering

\caption{Merging algorithm comparison using merging and attention timings. Bold latency values mark the best results among merging methods.}
\label{tab:merging_ablation}
\renewcommand{\arraystretch}{1.4}
\resizebox{\textwidth}{!}{%
\begin{tabular}{lcccccc}
\toprule
\textbf{Method} & \textbf{Merging (ms)} & \textbf{Attention (ms)} & \textbf{Total (ms)} & \textbf{L2 Error} & \textbf{CTA Cost ($10^9$)} & \textbf{Density (\%)} \\
\midrule
No Merging & -- & 69.28 & 69.36 & 0.674 & 12.79 & 19.88 \\
Deterministic Greedy & 3294.86 & 62.60 & 3357.47 & \textbf{0.627} & \textbf{11.01} & 22.27 \\
Weighted Overlap & 987.83 & 72.26 & 1060.09 & 0.654 & 13.74 & 22.56 \\
\rowcolor{mygold}
Parallel Agglomerative (Ours) & \textbf{257.06} & \textbf{62.38} & \textbf{319.44} & \textbf{0.627} & 11.23 & 22.46 \\
\bottomrule
\end{tabular}
}
\end{table}
}

\newcommand{\tabShortDenoising}{
\begin{table}[!htbp]
\centering
\tiny
\caption{Evaluation of distilled HunyuanVideo-1.5 with image conditioning on H200. Speedup is relative to full attention. OOM denotes out of memory.}
\label{tab:hunyuanvideo15}
\renewcommand{\arraystretch}{1.25}
\resizebox{0.9\textwidth}{!}{%
\begin{tabular}{lccccc}
\toprule
\multicolumn{1}{c}{\multirow{2}{*}{\textbf{Method}}} & \multicolumn{3}{c}{\textbf{Quality}} & \multicolumn{2}{c}{\textbf{Efficiency}} \\
\cmidrule(lr){2-4}\cmidrule(lr){5-6}
 & \textbf{PSNR} $\uparrow$ & \textbf{SSIM} $\uparrow$ & \textbf{LPIPS} $\downarrow$ & \textbf{Density} $\downarrow$ & \textbf{Speedup} $\uparrow$ \\
\midrule
\rowcolor{neutralgray}
\multicolumn{6}{c}{\textbf{HunyuanVideo-1.5} (\textit{Image-to-Video, 720p, Distilled, 12 Steps})} \\
\addlinespace[2pt]
Full Attention & -- & -- & -- & 100\% & $1.00\times$ \\
\addlinespace[-1pt]
\cmidrule(lr){1-6}
\addlinespace[-1pt]
SpargeAttn & 24.64 & 0.848 & 0.133 & 30.40\% & 1.94$\times$ \\
SVG (OOM) & -- & -- & -- & -- & -- \\
SVG2 & 25.73 & 0.879 & 0.110 & 23.38\% & 1.89$\times$ \\
\rowcolor{mygold}
\algnameshort & \textbf{27.58} & \textbf{0.914} & \textbf{0.083} & 28.91\% & 2.00$\times$ \\
\rowcolor{mygold}
\algnameshort-Flash & 26.70 & 0.897 & 0.098 & 25.75\% & \textbf{2.10$\times$} \\
\bottomrule
\end{tabular}
}
\end{table}
}

\newcommand{\tabTradeoffSweep}{
\begin{table}[!htbp]
\centering
\scriptsize
\caption{Tradeoff between quality and latency for SVG2 and \algnameshort~using HunyuanVideo on an H200 GPU. Speedup is relative to full attention. \textbf{Bold} denotes the best result within each sweep.}
\label{tab:tradeoff_sweep}
\renewcommand{\arraystretch}{1.25}
\resizebox{1\textwidth}{!}{%
\begin{tabular}{lcccccc}
\toprule
\multicolumn{2}{c}{} & \multicolumn{3}{c}{\textbf{Quality}} & \multicolumn{2}{c}{\textbf{Efficiency}} \\
\cmidrule(lr){3-5}\cmidrule(lr){6-7}
\textbf{Method} & \textbf{min\_kc} & \textbf{PSNR} $\uparrow$ & \textbf{SSIM} $\uparrow$ & \textbf{LPIPS} $\downarrow$ & \textbf{Density} $\downarrow$ & \textbf{Speedup} $\uparrow$ \\
\midrule
Full Attention & -- & -- & -- & -- & 100\% & $1.00\times$ \\
\addlinespace[-1pt]
\cmidrule(lr){1-7}
\addlinespace[-1pt]
SVG2 & 0.05 & 30.76 & 0.935 & 0.068 & 23.82\% & \textbf{2.19$\times$} \\
SVG2 & 0.0625 & 31.67 & 0.943 & 0.057 & 24.33\% & 2.18$\times$ \\
SVG2 & 0.075 & 32.44 & 0.950 & 0.050 & 24.94\% & 2.15$\times$ \\
SVG2 & 0.0875 & 32.79 & 0.953 & 0.047 & 25.54\% & 2.13$\times$ \\
SVG2 & 0.1 & \textbf{33.11} & \textbf{0.955} & \textbf{0.044} & 26.22\% & 2.10$\times$ \\
\cmidrule(lr){1-7}
\rowcolor{mygold}
\algnameshort-Flash & 0.1 & 33.71 & 0.961 & 0.039 & 28.31\% & \textbf{2.21$\times$} \\
\algnameshort & 0.1125 & 33.83 & 0.961 & 0.038 & 29.00\% & 2.19$\times$ \\
\algnameshort & 0.125 & 34.22 & 0.964 & 0.037 & 29.81\% & 2.14$\times$ \\
\algnameshort & 0.1375 & 34.17 & 0.963 & 0.037 & 30.54\% & 2.14$\times$ \\
\rowcolor{mygold}
\algnameshort & 0.15 & \textbf{34.73} & \textbf{0.967} & \textbf{0.034} & 31.33\% & 2.11$\times$ \\
\bottomrule
\end{tabular}
}
\end{table}
}

\newcommand{\algoLocalWindowClustering}{
\begin{algorithm}[t]
\caption{3D Local-Window Clustering in Triton}
\label{alg:videoslic}
\begin{algorithmic}[1]
\setlength{\itemsep}{3pt}
\State {\bf Inputs:} token tensor $X \in \mathbb{R}^{T \times H \times W \times D}$, grid size $(k_t, k_h, k_w)$, local-window size $(w_t, w_h, w_w)$, iterations $N_{\mathrm{iter}}$
\State {\bf Output:} assignments $\{a_\ell\}_{\ell=1}^{N}$, centroids $\{\mu_u\}_{u \in \mathcal{G}}$
\State $K_t \gets \lceil T / k_t \rceil,\; K_h \gets \lceil H / k_h \rceil,\; K_w \gets \lceil W / k_w \rceil$
\State $w_t \gets \min(w_t, K_t),\; w_h \gets \min(w_h, K_h),\; w_w \gets \min(w_w, K_w)$
\State $\mathcal{G} \gets [0, K_t\!-\!1] \times [0, K_h\!-\!1] \times [0, K_w\!-\!1]$
\State For token coordinate $\ell=(t_\ell,h_\ell,w_\ell)$, let
\Statex \hspace{\algorithmicindent}$g(\ell) \gets \left(\left\lfloor \frac{t_\ell}{k_t} \right\rfloor, \left\lfloor \frac{h_\ell}{k_h} \right\rfloor, \left\lfloor \frac{w_\ell}{k_w} \right\rfloor\right),\quad x_\ell \gets X_{t_\ell,h_\ell,w_\ell,:}$
\For{each grid $u=(u_t,u_h,u_w) \in \mathcal{G}$}
    \State $c_t \gets \min(u_t k_t + \lfloor k_t/2 \rfloor,\, T-1)$
    \State $c_h \gets \min(u_h k_h + \lfloor k_h/2 \rfloor,\, H-1)$
    \State $c_w \gets \min(u_w k_w + \lfloor k_w/2 \rfloor,\, W-1)$
    \State $\mu_u \gets X_{c_t,c_h,c_w,:}$
\EndFor
\For{$\eta = 1$ to $N_{\mathrm{iter}}$}
    \State $z_u \gets 0 \in \mathbb{R}^{D},\; n_u \gets 0 \qquad \forall u \in \mathcal{G}$
    \For{each grid $u=(u_t,u_h,u_w) \in \mathcal{G}$ {\bf in parallel}}
        \State $\bar{u}_t \gets \mathrm{clip}(u_t - \lfloor w_t/2 \rfloor,\, 0,\, K_t-w_t)$
        \State $\bar{u}_h \gets \mathrm{clip}(u_h - \lfloor w_h/2 \rfloor,\, 0,\, K_h-w_h)$
        \State $\bar{u}_w \gets \mathrm{clip}(u_w - \lfloor w_w/2 \rfloor,\, 0,\, K_w-w_w)$
        \State $\mathcal{N}(u) \gets \{(\bar{u}_t+\delta_t,\bar{u}_h+\delta_h,\bar{u}_w+\delta_w) \mid \delta_t \in [0,w_t),\, \delta_h \in [0,w_h),\, \delta_w \in [0,w_w)\}$
        \For{each token coordinate $\ell=(t_\ell,h_\ell,w_\ell)$ such that $g(\ell)=u$}
            \State $a_\ell \gets \arg\min_{v \in \mathcal{N}(u)} \|x_\ell - \mu_v\|_2^2$
            \State $z_{a_\ell} \gets z_{a_\ell} + x_\ell,\quad n_{a_\ell} \gets n_{a_\ell} + 1$ \Comment{atomic accumulation}
        \EndFor
    \EndFor
    \For{each grid $u \in \mathcal{G}$}
        \State $\mu_u \gets
        \begin{cases}
            z_u / n_u, & n_u > 0,\\
            \mu_u, & n_u = 0
        \end{cases}$
    \EndFor
\EndFor
\State \Return $\{a_\ell\}_{\ell=1}^{N}, \{\mu_u\}_{u \in \mathcal{G}}$
\end{algorithmic}
\end{algorithm}
}

\newcommand{\algoClusterMerging}{
\begin{algorithm}[t]
\caption{Parallel Agglomerative Merging for Hardware Cost Minimization}
\label{alg:greedy_agglomerative}
\begin{algorithmic}[1]
\setlength{\itemsep}{1pt}
\State {\bf Inputs:} query cluster indices $\mathcal{I}_q=\{1,\dots,N_q\}$, sizes $\{s_i^q\}$, $\{s_j^k\}$, mask $M=[m_{ij}]$, tile sizes $\tau_q,\tau_k$, rounds $R$
\State {\bf Output:} a disjoint partition $\mathcal{P}$ of $\mathcal{I}_q$
\State $\mathcal{P} \gets \big\{\{i\} \mid i \in \mathcal{I}_q\big\}$ \Comment{Initialize every query cluster as a singleton group}
\State Initialize $\chi_j(\{i\}) \gets m_{ij}$ for all $i \in \mathcal{I}_q$ and $j = 1, \dots, N_k$
\For{$r=1$ to $R$}
    \State Randomly bipartition the active groups in $\mathcal{P}$ into $\mathcal{A}$ and $\mathcal{B}$
    \For{each $A \in \mathcal{A}$}
        \For{each $B \in \mathcal{B}$}
            \State $s^q(A \cup B) \gets s^q(A) + s^q(B)$ \Comment{Sum query sizes}
            \State $\chi_j(A \cup B) \gets \max(\chi_j(A), \chi_j(B))$ for $j = 1, \dots, N_k$ \Comment{Logical OR}
            \State $C(A \cup B) \gets \left( \tau_q \left\lceil \frac{s^q(A \cup B)}{\tau_q} \right\rceil \right) \left( \tau_k \left\lceil \frac{\sum_{j=1}^{N_k} \chi_j(A \cup B) s_j^k}{\tau_k} \right\rceil \right)$
            \State $\Delta C(A, B) \gets C(A \cup B) - \big(C(A) + C(B)\big)$
        \EndFor
        \State $\pi(A) \gets \arg\min_{B \in \mathcal{B}} \Delta C(A,B)$
    \EndFor
    \State $\mathcal{M} \gets \{(A,\pi(A)) \mid \Delta C(A,\pi(A)) < 0 \text{ and } A = \arg\min_{A' : \pi(A')=\pi(A)} \Delta C(A',\pi(A))\}$
    \State $\mathcal{R} \gets \bigcup_{(A,B)\in\mathcal{M}} \{A,B\}$
    \State $\mathcal{P} \gets \left(\mathcal{P} \setminus \mathcal{R}\right) \cup \left\{A \cup B \mid (A,B)\in \mathcal{M}\right\}$ \Comment{Execute merges in parallel}
\EndFor
\State \Return $\mathcal{P}$
\end{algorithmic}
\end{algorithm}
}

\usepackage{amsmath,amsfonts,bbm,bm}

\newcommand{\inner}[1]{\left \langle {#1} \right \rangle}
\newcommand{\bigo}{\mathcal{O}}

\def\0{{\bm 0}}

\newcommand{\daggerthanks}[1]{\setcounter{footnote}{1}\thanks{#1}}
\newcommand{\daggerfootnotemark}{\footnotemark[2]}

\title{\algname: Efficient Sparse Attention  \\ with Spatio-Temporal Clustering for Video Diffusion}
\author{Dongyeun Lee$^{1}$ \quad Amir Zandieh$^{2}$ \quad Vahab Mirrokni$^{2}$ \quad Junmo Kim$^{1}$\daggerthanks{Corresponding Authors.} \quad Insu Han$^{1}$\daggerfootnotemark \\\\
$^{1}$KAIST \qquad $^{2}$Google Research
}
\date{June, 2026}

\begin{document}

\maketitle

\figTeaser

\begin{abstract}
Video Diffusion Transformers (VDiTs) have demonstrated significant capabilities in high-fidelity video generation. However, their ability to produce long-duration videos is fundamentally constrained by the quadratic complexity of the self-attention mechanism. Recent clustering-based sparse attention methods improve the quality-speed trade-off by grouping semantically similar tokens, but their practical efficiency remains limited by two bottlenecks: substantial clustering overhead and low CTA utilization caused by irregular cluster-induced blocks. We propose \algname~(\algnameshort), a training-free sparse attention framework that addresses both bottlenecks jointly. To reduce clustering overhead, we introduce 3D local-window clustering, which exploits the spatio-temporal locality of video tokens to restrict centroid search to fixed local neighborhoods, and implement it with a custom Triton kernel for efficient execution. We further propose a hybrid clustering strategy that performs full clustering only at anchor steps and updates only subset tokens at intermediate steps, leveraging the temporal stability of cluster assignments across denoising steps. To improve CTA utilization, we present hardware-aware cluster merging that minimizes CTA-aligned execution cost through parallel agglomerative merging, improving block density and approximation fidelity by utilizing idle tile capacity. Together, these components reduce clustering overhead, avoid redundant updates, and better align sparse attention with the fixed tile structure of modern GPU kernels. Experiments on video generation show that \algnameshort~advances the Pareto frontier of training free sparse attention, achieving higher fidelity than SVG2 at comparable latency and a $2.13\times$ speedup over full attention on HunyuanVideo, or $2.23\times$ with its faster variant \algnameshort-Flash.
\end{abstract}

\section{Introduction}
\label{sec:intro}
Diffusion Transformers (DiTs)~\cite{peebles2023scalable} have recently emerged as a powerful backbone for high-fidelity video generation~\cite{googleveo31,openaisora,lightricksltx2,yang2024cogvideox, kong2024hunyuanvideo,wan2025wan, zheng2024open,arkhipkin2025kandinsky}.
However, practical deployment remains severely constrained by spatio-temporal self-attention, which jointly attends over all tokens across frames and spatial locations, resulting in extremely long token sequences and quadratic complexity.
To address this bottleneck, various prior works \cite{cai2025mixture,zhang2025spargeattention,wu2026vmoba,xia2025training,zhang2025fast,sun2025vorta,xi2025sparse} construct block-sparse masks over the original token layout, using blocks aligned with Cooperative Thread Arrays (CTAs), the tile-level execution units of FlashAttention kernels \cite{dao2022flashattention}.
these approaches treat all tokens in a block as a coarse unit and fail to capture fine-grained token-level relevance, often compromising visual quality.
To overcome this limitation, recent studies \cite{han2024hyperattention,yang2025sparse,zhou2026svg} have introduced clustering-based sparse attention, which cluster tokens by fine-grained relevance and permute them so that similar tokens become contiguous, enabling sparse attention to be computed over clustered token groups, which are then processed as blocks by the attention kernel.
These clustering-based methods generally offer a better quality-speed trade-off than sparse attention over the original token layout.

Despite this progress, existing clustering-based approaches still face two key limitations that constrain their practical speedups.
First, existing methods incur substantial clustering overhead by failing to fully exploit spatio-temporal locality and temporal stability.
Prior methods often rely on generic global procedures such as \textit{k}-means that compare every token against every centroid without exploiting the spatio-temporal locality of video tokens, making clustering itself a major bottleneck even with highly optimized kernels.
They also overlook the temporal stability of token assignments across adjacent denoising steps, leading to redundant recomputation even though most assignments remain unchanged over time.
Second, cluster-induced blocks often lead to low thread utilization and substantial efficiency loss.
As noted above, clustering-based sparse attention uses token clusters as logical attention blocks. In practice, these cluster-induced blocks are executed after padding to fixed CTA tile sizes, as illustrated in \cref{fig:cluster_merging}(a).
Because cluster sizes are often misaligned with the CTA tile shape of attention kernels~\cite{dao2022flashattention,dong2024flex,ye2025flashinfer}, threads assigned to the padded region outside the true cluster extent remain underutilized, preventing theoretical sparsity from fully translating into actual speedup.
Together, these limitations motivate a method that exploits spatio-temporal redundancy and temporal stability while improving the execution efficiency of sparse attention on modern GPUs.

To this end, we present \algname (\algnameshort), an efficient sparse attention framework for video diffusion that addresses both clustering overhead and low thread utilization.
To reduce clustering overhead, we first introduce \textit{3D local-window clustering}, which restricts centroid search to local neighborhoods by exploiting the spatio-temporal locality of video tokens. Compared with generic global clustering (e.g. \textit{k}-means), it converges faster, has lower complexity, and is implemented with a custom Triton kernel for efficient execution.
We further introduce a \textit{hybrid clustering} strategy that leverages the temporal stability of token assignments across denoising steps by performing full clustering only at anchor steps and clustering only a subset of tokens at intermediate steps.
To address low thread utilization from irregular cluster blocks, we further propose \textit{hardware-aware cluster merging}, which formulates the tiled execution cost of sparse attention as a hardware cost minimization problem and approximately solves it with efficient parallel agglomerative merging. By merging clusters to reduce hardware cost, this step improves CTA utilization and latency, while using unused tile capacity to accommodate a more accurate attention approximation at no extra hardware cost.
Together, these components enable \algnameshort~to advance the Pareto frontier of training free sparse attention for video generation, achieving higher fidelity than SVG2 at comparable latency and a $2.13\times$ speedup over full attention on HunyuanVideo~\cite{kong2024hunyuanvideo}, or $2.23\times$ with its faster variant \algnameshort-Flash.

\figClusterMerging

\section{Preliminaries}

Let $Q,K,V\in\mathbb{R}^{n\times d}$ denote the query, key, and value matrices of a multi-head self-attention with $h$ heads, where $n$ is the number of tokens and $d$ is the per-head dimension.
\citet{yang2025sparse} proposed SVG2, a clustering-based sparse attention approximation for VDiT, which applies \textit{k}-means clustering to query and key embeddings respectively, and consequently each one is partitioned into $N_q$ and $N_k$ clusters.
Using the corresponding permutation matrices $\Pi^q$ and $\Pi^k$, the reordered tensors are
\begin{align}
Q' = \Pi^q Q, \qquad K' = \Pi^k K, \qquad V' = \Pi^k V,
\end{align}
which place semantically similar tokens contiguously in memory.
This produces cluster-induced blocks for block-sparse attention.
Let $s_i^q$ and $s_j^k$ denote the query and key cluster sizes, and let $\mu_i^q$ and $\mu_j^k$ denote their corresponding centroids.
A common approximation is to estimate cluster-pair importance using centroid-level scores
\begin{equation}
S_{ij} = \frac{ \inner{\mu_i^q, \mu_j^k}}{\sqrt{d}}, \qquad
P'_{ij} = \frac{s_j^k \exp(S_{ij})}{\sum_{o=1}^{N_k} s_o^k \exp(S_{io})}.
\label{eq:weighted-attn-scores}
\end{equation}
where $\mu_i^q$ and $\mu_j^k$ are the centroids of the $i$-th query cluster and $j$-th key cluster, respectively.
The cluster-level probabilities are converted into a sparse mask by a row-wise Top-$p$ rule with threshold $p\in(0,1]$.
For each query cluster $i$, let $\mathrm{TopP}(P'_{i,:};p)$ denote the smallest set of key-cluster indices, taken in descending order of $P'_{i,:}$, whose cumulative probability is at least $p$.
We define
\begin{equation}
\mathcal{A}_i = \mathrm{TopP}(P'_{i,:};p), \qquad
m_{ij} = \mathbbm{1}\!\left[j \in \mathcal{A}_i\right].
\label{eq:cluster-top-p-mask}
\end{equation}
This yields a cluster-level sparse mask $M=[m_{ij}] \in \{0,1\}^{N_q \times N_k}$.
The selected key clusters define the block-sparse attention computation:
\begin{equation}
O_i'=\mathrm{Attn}\!\left(Q_i',\{K_j',V_j'\}_{j\in\mathcal{A}_i}\right),
\qquad
O={\Pi^q}^{-1}O'.
\label{eq:cluster-block-sparse-attn}
\end{equation}
Here, $\mathrm{Attn}$ denotes standard scaled dot-product attention restricted to the selected key/value clusters.
However, the resulting cluster-induced blocks often have sizes misaligned with fixed CTA tile shapes, causing thread underutilization.
Our method follows this clustering-based sparse attention pipeline, while reducing clustering overhead and improving hardware efficiency through cluster merging.

\section{Motivation}
\label{sec:motivation}
Existing clustering-based sparse attention improves the quality-speed tradeoff by grouping semantically similar tokens, but its realized speedup remains limited by clustering overhead and low utilization of fixed GPU tiles.
We analyze three observations behind these limits. Video tokens are locally coherent in space and time, cluster assignments are stable across denoising steps, and token sparsity does not always translate into efficient CTA execution.

\noindent\textbf{Spatio-Temporal Locality of Video Tokens.}
Videos exhibit substantial redundancy in space and time: nearby regions are often similar, and consecutive frames change gradually.
To verify whether this property also holds in diffusion latent space, we cluster tokens with \textit{k}-means and visualize the assignments by projecting 128-dimensional head features into RGB space using Principal Component Analysis (PCA).
As shown in \cref{fig:motivation-cluster}, spatially and temporally adjacent latent tokens tend to fall into the same or similar clusters.
This observation suggests that meaningful assignments can often be identified from local neighborhoods alone, which motivates replacing global clustering with a local alternative that exploits the intrinsic 3D structure of video tokens and reduces clustering overhead.

\figMotivationCombined
\noindent\textbf{Temporal Stability across Denoising Steps.}
Recent works~\cite{Liu_2025_CVPR,wimbauer2024cache,zhao2025realtime,lv2025fastercache} show that diffusion latent features are stable across adjacent denoising steps.
We find that this stability extends to per-token cluster assignments.
As shown in \cref{fig:motivation-change}, cluster assignments remain stable across most layers and timesteps. Beyond denoising step 10, only 25.8\% of query assignments and 27.6\% of key assignments change on average.
Moreover, to examine which tokens remain stable, we use Lowe's ratio~\cite{lowe2004distinctive}, defined as the ratio between the nearest- and second-nearest-centroid distances.
A larger ratio indicates that the token lies near a cluster boundary and has an unstable assignment.
As shown in \cref{fig:motivation-lowe}, tokens that later change clusters tend to have higher Lowe's ratios, making it a useful signal for subset selection.
This motivates a hybrid update rule with infrequent full clustering and partial updates of only a subset of tokens at intermediate steps.

\noindent\textbf{From Theoretical Density to Hardware Efficiency.}
In clustering-based sparse attention, low theoretical token density is not sufficient to yield end-to-end speedup. Attention is executed on GPUs at the granularity of cooperative thread arrays (CTAs), whereas these methods construct sparse blocks from token clusters. These cluster-induced blocks are generally misaligned with the fixed CTA tile sizes of FlashAttention kernels (e.g., $128$ queries and $96$ keys on Hopper), resulting in thread underutilization. As detailed in \cref{fig:speed_density_comparison}, a pattern with $39.7\%$ theoretical density still incurs $50.4\%$ CTA density. This motivates a hardware-aware post-processing step that better aligns cluster-induced blocks with CTA tile sizes while preserving attention approximation quality.

\section{\algname}
\label{sec:method}

\subsection{3D Local-Window Clustering}
\label{sec:clustering}
As discussed in \cref{sec:motivation}, video tokens exhibit strong spatio-temporal redundancy, making an expensive global centroid search often unnecessary.
Building on this observation, we propose an efficient 3D local-window clustering, which restricts clustering to local neighborhoods instead of performing a global centroid search.
Inspired by over-segmentation methods that partition images into a regular grid~\cite{zitnick2007stereo,achanta2012slic}, we extend this principle to a 3D spatio-temporal token lattice, a design visually supported by the spatially coherent cluster assignments in \cref{fig:motivation-cluster}.
As illustrated in \cref{fig:lw_clustering_kernel_speed_lw_clustering}, we partition the spatio-temporal token lattice into regular 3D grids of size $(k_t, k_h, k_w)$, so that each token belongs to exactly one grid.
The number of centroids is identical to the number of grids. Each centroid is initialized from the token at the center of a grid, and the resulting centroids preserve the same 3D topology as the underlying grid layout.
Let $x_\ell \in \mathbb{R}^d$ denote the embedding of token $\ell$, and let $\mu_u \in \mathbb{R}^d$ denote the centroid associated with grid $u$.
For a token $\ell$ at coordinate $(t_\ell, h_\ell, w_\ell)$, let $g(\ell)$ denote the grid containing token $\ell$, and define its local assignment as
\begin{align}
g(\ell)=\left(\left\lfloor \frac{t_\ell}{k_t}\right\rfloor,\left\lfloor \frac{h_\ell}{k_h}\right\rfloor,\left\lfloor \frac{w_\ell}{k_w}\right\rfloor\right),
\qquad
a_\ell=\mathop{\mathrm{argmin}}_{u\in\mathcal{N}(g(\ell))}\|x_\ell-\mu_u\|_2^2,
\end{align}
where $\mathcal{N}(g(\ell))$ denotes the set of centroids in a local 3D window of size $(w_t, w_h, w_w)$ around the grid $g(\ell)$.
This restricts the search to a local spatio-temporal neighborhood and avoids redundant comparisons against distant centroids.
The centroids are then updated by averaging the embeddings of their assigned tokens, while the underlying grid partition remains fixed throughout the iterations.

This avoids the usual quality-cost tradeoff in generic clustering, where more centroids improve quality but increase assignment cost.
In methods such as \textit{k}-means clustering, each of the $n$ tokens must be compared against all $N_c$ centroids, resulting in $\bigo(n\,N_c)$ complexity.
In video diffusion, where $n$ is already large, this quickly becomes a bottleneck.
By contrast, 3D local-window clustering restricts the search volume of each token to a fixed local window with volume $w_t w_h w_w$, yielding $\bigo(n\,w_t w_h w_w)$ assignment cost independent of the total number of centroids.
For HunyuanVideo at 720p, with $n=118.8\mathrm{K}$, $N_c=1215$, and $(w_t,w_h,w_w)=(3,8,8)$, this yields a $6.3\times$ reduction in the per-token search space.
This decouples clustering granularity from assignment overhead, allowing a larger centroid set to improve clustering quality without increasing assignment cost.

\noindent\textbf{Efficient GPU Kernel Implementation.}
We implement 3D local-window clustering with a custom Triton kernel that performs local assignment and centroid accumulation directly on the native tensor layout.
This avoids explicitly constructing overlapping 3D local grids from tokens stored in flattened $T\!\times\!H\!\times\!W$ layout, which would incur high memory overhead.
Our design is inspired by implicit GEMM \cite{Thakkar_CUTLASS_2023}. Following the same principle, the kernel forms 3D neighborhood tiles on the fly through index arithmetic rather than explicitly materializing overlapping local grids, and uses them for distance evaluation and centroid-statistics accumulation.
Compared with a naive PyTorch implementation that explicitly rearranges tokens into overlapping local grids, our Triton kernel achieves up to a $411\times$ speedup (\cref{fig:lw_clustering_kernel_speed_kernel}). Implementation details are provided in \cref{sup:algo}.

\figLocalWindowClusteringKernelSpeed

\subsection{Hybrid Clustering across Denoising Steps}
As discussed in \cref{sec:motivation}, most token assignments remain stable across adjacent denoising steps, and Lowe's ratio \cite{lowe2004distinctive} provides a useful signal for identifying the subset most likely to change. Based on this observation, we adopt a hybrid clustering strategy that performs full-token 3D local-window clustering only at anchor steps and updates only a subset of tokens at intermediate steps, as illustrated in \cref{fig:motivation-lowe}.
Specifically, we use Lowe's ratio,
\begin{align}
\rho_\ell = \frac{d_{\ell,1}}{d_{\ell,2}}, \qquad
d_{\ell,1} = \min_{u} \|x_\ell - \mu_u\|_2, \qquad
d_{\ell,2} = \min_{u \neq u_\ell^\star} \|x_\ell - \mu_u\|_2,
\label{eq:lowes_ratio}
\end{align}
where $u_\ell^\star = \arg\min_{u} \|x_\ell - \mu_u\|_2$ denotes the nearest centroid of token $\ell$.

At each first denoising step, we perform full-token 3D local-window clustering, compute the value $\rho_\ell$ for all tokens, and cache the indices of a subset of tokens with the largest $\rho_\ell$ values.
At subsequent intermediate steps, we update only this subset by performing a standard \textit{k}-means reassignment against the current centroid set and then updating the corresponding centroids.
This substantially reduces computation, since most tokens are not revisited between anchor steps.
It also complements 3D local-window clustering with a global refinement step: whereas local-window clustering restricts each token to nearby centroids, subset refinement allows unstable tokens to be reassigned using the full centroid set.
In this way, the hybrid strategy combines low overhead with periodic global correction.

\subsection{Hardware-Aware Cluster Merging}
As discussed in \cref{sec:motivation}, cluster sizes are often misaligned with the fixed CTA tile shapes of attention kernels, leading to thread underutilization.
To address this issue, we formulate the hardware-efficiency objective of sparse attention as a cost minimization problem, and then present a parallel agglomerative heuristic for approximately solving it.

\noindent\textbf{Hardware Cost Minimization.}
Since the softmax operation is applied independently to each row of the attention matrix, each query cluster constitutes a single block row and is executed independently.
For each row, efficient modern sparse attention kernel \cite{ye2025flashinfer} packs the selected key cluster blocks into contiguous memory, enabling efficient computation in units of CTA tiles while avoiding thread underutilization.
Accordingly, the hardware cost of a single query cluster $i$ can be written as
\begin{align}
    C(i) = \left(\tau_q \left\lceil \frac{s_i^q}{\tau_q} \right\rceil \right)
    \left(\tau_k \left\lceil \frac{\sum_{j=1}^{N_k} m_{ij} s_j^k}{\tau_k} \right\rceil \right),
    \label{eq:mergeq-single}
\end{align}
where $N_q$ and $N_k$ denote the numbers of query and key clusters, $\tau_q$ and $\tau_k$ are the CTA tile sizes, $M=[m_{ij}] \in \{0,1\}^{N_q \times N_k}$ is the cluster-level sparse mask, and $s_i^q$ and $s_j^k$ are the sizes of query cluster $i$ and key cluster $j$, respectively.
In \cref{eq:mergeq-single}, the two terms are the CTA-aligned sizes of the query block and packed key blocks.
Thus, $C(i)$ measures tiled work rather than token sparsity.

\noindent\textbf{Parallel Agglomerative Merging.}
A straightforward way to reduce this cost is to fill underutilized CTA tiles by merging query clusters with compatible key selections.
We extend the row-wise cost in \cref{eq:mergeq-single} from a single query cluster to a merged group. When multiple query clusters are merged, their query sizes add up, while the selected key clusters are given by the union of their row masks. For a group $S \subseteq \mathcal{I}_q$ with $\mathcal{I}_q=\{1,\dots,N_q\}$, we define $s^q(S) = \sum_{i \in S} s_i^q$ and $\chi_j(S) = \max_{i \in S} m_{ij}$. Then,
the corresponding execution cost becomes
\begin{align}
    C(S)
    =
    \left(\tau_q \left\lceil \frac{s^q(S)}{\tau_q} \right\rceil \right)
    \left(\tau_k \left\lceil \frac{\sum_{j=1}^{N_k} \chi_j(S) s_j^k}{\tau_k} \right\rceil \right).
    \label{eq:mergeq-set}
\end{align}
This induces the following partition problem over query clusters:
\begin{align}
    \min_{\mathcal{P}  \text{: Partition of } \mathcal{I}_q} \sum_{S \in \mathcal{P}} C(S).
    \label{eq:mergeq-obj}
\end{align}
Since \cref{eq:mergeq-obj} is a combinatorial set partitioning problem with a non-additive block cost, approximation strategies are possible including exact combinatorial solvers and sequential greedy agglomerative merging.
In our setting, however, cluster merging is useful only if its overhead remains negligible relative to the hardware cost it saves.
This makes both exact optimization and sequential greedy updates unattractive in practice.
Thus, our goal is to avoid sequential single-merge updates and instead use a parallel merge scheme.
To this end, we propose a parallel agglomerative algorithm: in each round, we randomly bipartition the active groups, evaluate only cross-partition candidates, and merge multiple disjoint cost-reducing pairs in parallel.
Disjointness ensures that no group participates in more than one accepted merge and can therefore be applied in parallel without conflict. More formally, for two disjoint groups $A$ and $B$, define the merge gain
\begin{align}
\Delta C(A, B) := C(A \cup B) - C(A) - C(B).
\end{align}
If $\Delta C(A, B) < 0$, merging $A$ and $B$ reduces hardware cost.
For each group in one subset, we select the group in the other subset with the smallest merge gain. If multiple groups select the same target, we keep only the best match, so the accepted pairs remain disjoint and can be executed in parallel. Repeating this procedure progressively builds larger groups over multiple rounds. Implementation details are provided in \cref{alg:greedy_agglomerative}.

Although this procedure does not guarantee the global optimum of \cref{eq:mergeq-obj}, it is highly parallelizable and directly targets the execution overhead caused by tile misalignment. Since merging takes the union of the selected clusters, it can introduce additional attention computations within idle tile capacity, improving approximation fidelity while reducing hardware cost.

\algoClusterMerging

\section{Experiments}
\label{sec:exp}
\noindent\textbf{Experimental Settings.}
We evaluate \algname~(\algnameshort)~ using Wan2.2-I2V/T2V-A-14B \cite{wan2025wan} and HunyuanVideo-T2V-13B \cite{kong2024hunyuanvideo} for video generation on a single NVIDIA H200 GPU.
We generate 81 and 129 frames at 720p resolution for Wan2.2 and HunyuanVideo, corresponding to token lengths of $n=75{,}600$ and $118{,}800$, respectively.
For Text-to-Video, we utilize the Penguin Benchmark\footnote{\url{https://github.com/Tencent-Hunyuan/HunyuanVideo/blob/main/assets/PenguinVideoBenchmark.csv}}, which provides optimized prompts derived from VBench \cite{huang2023vbench} dataset. For image-to-video, we employ center-cropped images from VBench at 720p. We randomly sample 50 instances for each task as in \cite{yang2025sparse}.
We compare against state-of-the-art training-free sparse attention baselines for video diffusion, including SpargeAttn~\cite{zhang2025spargeattention} and SVG~\cite{xi2025sparse}, which operate on the original token layout, and SVG2~\cite{yang2025sparse}, a clustering-based baseline.
We tune \algnameshort~for higher fidelity at latency comparable to SVG2 and reduce the minimum key cluster retention ratio (\texttt{min\_kc}) in its faster variant \algnameshort-Flash to maximize speed while retaining higher fidelity than SVG2. Detailed hyperparameters are provided in \cref{tab:hyperparams}.

\noindent\textbf{Metrics.} To measure fidelity in generated videos against the full-attention baseline, we employ LPIPS~\cite{zhang2018perceptual}, Structural Similarity Index Measure (SSIM) and PSNR. We also use scores in VBench~\cite{huang2023vbench} to evaluate overall video generation quality.
We additionally report an LLM-as-a-Judge score, inspired by recent LLM-based evaluation protocols \cite{zheng2023judging,chen2024mllm}, to capture complementary aspects such as subject consistency, aesthetics, and temporal flickering. We choose $\mathtt{gemini}$-$\mathtt{3}$-$\mathtt{flash}$ $\mathtt{preview}$~\cite{team2023gemini} and more details including evaluation prompt are provided in \cref{sup:llm_as_judge}.

\tabMain
\subsection{Results on Text/Image-to-Video Generation}
\cref{tab:main} reports the benchmarking results evaluating visual quality against inference efficiency across Wan 2.2 and HunyuanVideo.
Our main configuration, \algnameshort, achieves the best fidelity to full attention among sparse methods across the evaluated settings.
On HunyuanVideo text-to-video generation, it reaches 33.97 PSNR and 0.046 LPIPS, compared with 32.66 PSNR and 0.056 LPIPS for SVG2.
On Wan 2.2 image-to-video and text-to-video generation, it also gives the lowest perceptual degradation among sparse methods.
SpargeAttn obtains strong VBench scores in some settings, but its fidelity metrics are consistently worse than ours, indicating larger deviation from the full-attention output.
In contrast, under the LLM-as-a-Judge evaluation based on a recent vision-language model, \algnameshort~ achieves higher overall quality scores than the other sparse methods.
\algnameshort-Flash achieves speedups of $1.86\times$ on Wan2.2 image conditioned generation, $1.81\times$ on Wan2.2 text conditioned generation, and $2.23\times$ on HunyuanVideo.
Notably, our theoretical density is slightly higher than SVG2, both of ours achieve superior wall-clock speedups. This empirical finding validates our cluster merging strategy, showing that GPU-friendly block utilization can effectively overcome the operational bottlenecks of clustering-based sparse attention.

\subsection{Ablation Study}
\noindent\textbf{End-to-End Speedup.} As illustrated in \cref{fig:speed_latency_breakdown}, the latency breakdown reveals the source of our efficiency.
While SVG2 suffers from a heavy 83 s clustering overhead, our hybrid strategy with 3D local-window clustering requires only 26 s, accelerating the clustering stage by over $3.1\times$.
Furthermore, cluster merging improves hardware efficiency by aligning cluster block sizes with CTA tile sizes, with only 4 s of additional overhead.
Consequently, this combination of reduced clustering latency and efficient cluster merging enables \algnameshort-Flash to achieve up to a $2.23\times$ speedup over full attention, delivering the fastest inference among the evaluated methods, as shown in Table~\ref{tab:main}.

\noindent\textbf{Clustering Efficiency.}
To demonstrate the effectiveness of our 3D local-window clustering, we compare clustering error, which is closely correlated with the final output quality \cite{yang2025sparse}, against Flash \textit{k}-means \cite{yang2025sparse} across varying iterations in \cref{fig:speed_clustering_efficiency}.
Leveraging the spatio-temporal locality of video tokens, our 3D local-window clustering strategy proves highly effective and converges rapidly.
Our method requires only 10 iterations ($\sim$40 ms) to achieve a clustering error comparable to what \textit{k}-means reaches after 50 iterations ($>$250 ms). This achieves a remarkable $6.29\times$ latency reduction, effectively resolving the computational bottleneck of iterative token clustering.

\noindent\textbf{3D Local-Window Clustering Kernel Speed Evaluation.}
As shown in \cref{fig:lw_clustering_kernel_speed_kernel}, when evaluated across varying local window sizes, the custom Triton implementation of 3D local-window clustering consistently outperforms its PyTorch counterpart, achieving up to a $411\times$ speedup.
This gap is primarily a systems effect rather than an algorithmic one. The PyTorch baseline realizes the local search through a sequence of fine-grained tensor ops, incurring repeated kernel launches, global-memory round trips for intermediate tensors, and poor reuse across neighbor comparisons. In contrast, our Triton kernel keeps the computation on the native $T\!\times\!H\!\times\!W$ layout and fuses neighbor traversal, distance evaluation, Lowe's ratio computation, and centroid accumulation into a single pass, so each token feature is loaded once and consumed immediately. This substantially improves locality and arithmetic intensity while eliminating most launch overhead and redundant memory traffic.

\figLatency
\noindent\textbf{Hardware-Aware Cluster Merging.}
\cref{fig:speed_density_comparison} compares the theoretical token density and the actual CTA density of the attention kernel with and without cluster merging. Due to the misalignment between cluster size and thread-block tiling, underutilization is inevitable and leads to computational inefficiency. By integrating cluster merging, our approach accommodates a more accurate approximation of attention (theoretical density: $39.70\% \to 42.24\%$) within idle tile capacity, while reducing the actual CTA density from $50.42\%$ to $46.88\%$, corresponding to a $7.02\%$ relative reduction.

\section{Related Works}
\label{sec:related}
Existing methods are broadly divided into training-based \cite{zhang2025faster,zhang2025sla,zhang2026sla2,zhang2026spargeattention2,agarwal2026monarchrt,li2025radial,tan2025dsv,liang2026vmonarch} and training-free approaches~\cite{cai2025mixture,zhang2025spargeattention,wu2026vmoba,xia2025training,zhang2025fast,sun2025vorta,xi2025sparse,yang2025sparse,zhou2026svg}. Among training-free methods, several works \cite{cai2025mixture,zhang2025spargeattention,wu2026vmoba,xia2025training,zhang2025fast,sun2025vorta,xi2025sparse} construct block-sparse masks over the original token layout, which is hardware-friendly but often miss fine-grained token relevance. Clustering-based methods \cite{yang2025sparse,zhou2026svg} instead cluster and permute tokens to form cluster-induced blocks, better preserving semantic structure but remaining bottlenecked by iterative clustering and inefficient CTA utilization. Our method targets these two bottlenecks jointly.

Sparse Attention in LLMs has been extensively explored for long-context inference.
Several studies identify critical tokens or local attention sinks and retain them within constrained KV caches \cite{xiao2023streamingllm,zhang2023h2o}, estimate block- or page-level importance using proxy metrics to adaptively allocate attention budgets \cite{tang2024quest,jiang2024minference,lai2025flexprefill,xu2025xattention}, or reorder and hashe tokens to cluster relevant features efficiently \cite{han2024hyperattention,li2025mminference}.
However, these methods rely on 1D causal structure and target prefill or KV-cache efficiency, whereas video diffusion attention remains compute-bound over 3D spatiotemporal dependencies.

Orthogonal acceleration techniques include linear attention \cite{xie2024sana,chen2025sana}, high-compression auto-encoders \cite{chen2024deep}, quantization \cite{shang2023post,li2023q,huang2024tfmq,li2024svdquant,lee2025dmq,wu2024ptq4dit,zhao2024vidit,he2024efficientdm,sui2024bitsfusion,zhang2025sageattention,zhang2024sageattention2,zhang2025sageattention3}, caching \cite{Liu_2025_CVPR,wimbauer2024cache,liu2024faster,zhao2025realtime,lv2025fastercache,chen2025accelerating,ma2025model}, and distillation \cite{yin2024one,yin2024improved,chen2025sanasprint,zheng2025large}. Our work is complementary to these directions and focuses specifically on the quality-efficiency trade-off of sparse attention in video diffusion transformers.

\section{Conclusion}
\label{sec:conclusion}
We presented \algname~(\algnameshort), a training-free sparse attention framework that accelerates video diffusion by reducing clustering overhead and improving CTA utilization. It combines 3D local-window clustering, hybrid clustering strategy, and hardware-aware cluster merging to improve both algorithmic and kernel-level execution efficiency. Experiments on Wan2.2 and HunyuanVideo show that \algnameshort~advances the Pareto frontier of training free sparse attention, achieving higher fidelity than SVG2 at comparable latency and a $2.13\times$ speedup over full attention on HunyuanVideo, or $2.23\times$ with its faster variant \algnameshort-Flash. As a limitation, benefits are most pronounced for long videos where attention dominates runtime, and gains may be smaller for images or very short videos.

\section{Acknowledgements}
This work was supported by the National Research Foundation of Korea (NRF) grant funded by the Korea government (MSIT) (No. RS-2025-23523958, RS-2026-25470670) and the Institute of Information \& Communications Technology Planning \& Evaluation (IITP) grant funded by the Korea government (MSIT) (No. RS-2026-25617720, AI Star Fellowship Support Program).

\bibliographystyle{plainnat}
\bibliography{main}

\newpage
\appendix
\section{Implementation Details}
\label{sup:implementation}
\subsection{Hyperparameters}
\label{sup:hyperparam}
For hybrid clustering, we perform full clustering every two denoising steps and cache the top $15\%$ most ambiguous tokens, ranked by Lowe's ratio, for intermediate-step updates. For hardware-aware cluster merging, we use $R=3$ rounds of parallel agglomerative merging.
Table~\ref{tab:hyperparams} summarizes the hyperparameters used for all methods and evaluation settings. To ensure a fair comparison, we apply the same warmup strategy to every sparse baseline. Below we clarify the meaning of each table entry.

\noindent \textbf{Time warmup / Layer warmup.} Time warmup uses full attention during the first few denoising timesteps, and layer warmup uses full attention in the first few transformer layers. Each entry is reported as warmup length / total number of timesteps or layers, with stage-specific values listed separately when applicable.
To ensure a fair comparison, we apply identical warmup strategies across all evaluated methods.

\noindent \textbf{Q/K grid size.} The 3D temporal-height-width grid used to initialize the query and key centroids.

\noindent \textbf{Q/K window size.} The corresponding local search window used by 3D local-window clustering.

\noindent \textbf{$N_q$, $N_k$.} The numbers of query and key centroids, respectively.

\noindent \textbf{Top-\textit{p}.} Following SVG2, the cumulative centroid-selection threshold used to build the cluster-level sparse map. For each query centroid, we compute centroid-to-centroid attention scores, apply a size-weighted softmax over key centroids, and keep the highest-probability key centroids until their cumulative probability reaches $p$.

\noindent \textbf{Min-\textit{kc}.} The minimum retained key-cluster ratio for each query row. We always preserve at least this fraction of key clusters even if the Top-\textit{p} threshold is reached earlier.

\noindent \textbf{Density.} A parameter for SpargeAttn and SVG that controls their target block density.

\tabConfig
\subsection{Algorithms}
\label{sup:algo}
\noindent \textbf{3D Local-Window Clustering Kernel.}
Algorithm~\ref{alg:videoslic} summarizes the core computation of our 3D local-window clustering kernel. Rather than materializing overlapping local windows, the kernel derives local neighborhoods by index arithmetic on the native $T \times H \times W$ layout, performs local nearest-centroid assignment, and accumulates centroid statistics for the subsequent update step.

\algoLocalWindowClustering

\clearpage

\section{Additional Quantitative Results}
\subsection{Tradeoff between Quality and Latency}
\label{sup:tradeoff}
We compare \algnameshort~and SVG2 across multiple sparsity settings to determine whether \algnameshort~improves generation fidelity at comparable total generation latency. We evaluate both methods using HunyuanVideo on an H200 GPU, generating 15 videos per configuration. We vary the minimum ratio of retained key clusters, denoted \texttt{min\_kc}, while keeping the Top $p$ threshold at $0.9$.

\cref{tab:tradeoff_sweep} shows that \algnameshort~achieves higher fidelity across the evaluated latency range. At the same $2.19\times$ speedup, it improves PSNR from $30.76$ to $33.83$. At approximately $2.1\times$ speedup, it achieves $34.73$ PSNR versus $33.11$ for SVG2, with corresponding improvements in SSIM and LPIPS.

\tabTradeoffSweep


\subsection{Component Ablation}
\label{sup:component_ablation}
We evaluate the incremental contributions of hybrid updates and cluster merging guided by CTA cost on top of clustering within 3D local windows. For each configuration, we generate 15 videos using HunyuanVideo on an H200 GPU. We start with local clustering alone, then enable hybrid updates, and finally add cluster merging to obtain the complete \algnameshort~configuration.

As shown in \cref{tab:component_ablation}, hybrid updates retain generation fidelity while providing the improvement in inference speed. PSNR increases from $34.44$ to $34.59$, and speedup increases from $2.05\times$ to $2.06\times$. Adding cluster merging further improves all three fidelity metrics and raises speedup to $2.11\times$. Notably, merging increases theoretical token density from $28.83\%$ to $31.33\%$ while reducing latency. This behavior is consistent with the CTA cost objective because additional attention pairs can occupy otherwise idle CTA capacity, allowing higher fidelity without a proportional increase in tiled work.

\tabComponentAblation

\subsection{Comparison of Merging Algorithms}
\label{sup:merging_ablation}
We compare algorithms for cluster merging to examine how the merging strategy affects CTA cost, approximation error, and computational overhead. We consider deterministic greedy merging, which applies the same CTA cost objective sequentially, and weighted overlap merging, which estimates merge compatibility from the overlap between masks of selected key clusters. Our method evaluates CTA cost while accepting multiple disjoint merges in parallel. We measure merging and attention latency separately to distinguish optimization overhead from attention execution time.

As shown in \cref{tab:merging_ablation}, parallel agglomerative merging attains a CTA cost close to deterministic greedy and matches its L2 error of $0.627$. Applying merges in parallel substantially reduces optimization overhead, with merging latency of $257.06$ ms compared with $3294.86$ ms for the sequential greedy procedure. Weighted overlap also reduces merging overhead relative to deterministic greedy, but increases CTA cost and attention latency relative to the unmerged pattern. This illustrates the limitation of using mask overlap as a proxy, since similar key selections do not necessarily yield efficient CTA alignment after merging. Our method combines direct CTA cost optimization with parallel execution, achieving the lowest attention and combined latency among the merging algorithms.

\tabMergingAblation


\subsection{Distilled Video Generation with Few Denoising Steps}
\label{sup:short_denoising}
We evaluate \algnameshort~on a distilled video diffusion model to test its effectiveness with a short denoising trajectory. We use distilled HunyuanVideo-1.5 to generate videos at 720p from input images with 12 denoising steps on H200.

\cref{tab:hunyuanvideo15} shows that \algnameshort~achieves the highest fidelity among the evaluated sparse methods, improving PSNR from SVG2's $25.73$ to $27.58$ and speedup from $1.89\times$ to $2.00\times$. \algnameshort-Flash further increases speedup to $2.10\times$, the fastest evaluated configuration, while maintaining better PSNR, SSIM, and LPIPS than SVG2. SVG exceeds the available GPU memory in this setting.

\tabShortDenoising

\clearpage

\section{LLMs-as-a-Judge Results}
\label{sup:llm_as_judge}
\subsection{Evaluation Protocol}
Inspired by recent LLM-as-a-Judge evaluation approaches~\citep{zheng2023judging,chen2024mllm}, we use an LLM judge to complement automatic metrics with perceptual assessments of generated videos. We use a subset of VBench prompts as the text conditions and ask the judge to evaluate each generated video along three axes: \textit{subject consistency}, \textit{aesthetic quality}, and \textit{temporal flickering}. The judge assigns a score from 1 to 10 for each axis and returns a JSON object, from which we compute the overall LLM-as-a-Judge score as the average of the three scores. The detailed prompt is provided in \cref{subsec:eval_prompt}.

\tabLLMJudge

\subsection{Detailed Evaluation Result}
\cref{tab:video_results} reports the LLM-as-a-Judge results averaged over all evaluated samples. Overall, \algname~ achieves the best overall score among sparse methods across all settings, while Ours-Flash maintains competitive perceptual quality under a more aggressive speed-oriented configuration. These results suggest that our method preserves video quality beyond frame-level fidelity metrics.

\clearpage

\subsection{Evaluation Prompt}
\label{subsec:eval_prompt}
\begin{Verbatim}[breaklines=true, fontsize=\footnotesize]
1. System Role
- You are an expert Video Quality Assurance Engineer and Cinematographer. Your task is to evaluate an AI-generated video based on specific technical and artistic dimensions. You must provide objective, critical scores from 1 (Poor) to 10 (Excellent).

2. Evaluation Process
- Analyze the video frame-by-frame to detect inconsistencies.
- Assess the motion dynamics and temporal stability.
- Score each dimension based on the strict criteria provided.
- Calculate the final average.

3. Strict Evaluation Criteria

Dimension | Description & Scoring Guidelines

1. Subject Consistency | Does the main subject maintain its identity (features, clothing, shape) throughout the entire duration? Penalty for morphing.
2. Aesthestic Quality| General visual appeal, composition, lighting, and "cinematic" feel.
3. Temporal Flickering | Presence of high-frequency noise or "flashing" brightness/texture changes between frames.


4. Required Output Format

You MUST ONLY output a single, valid JSON object. Do NOT include any introductory text, analysis, or conversational response outside of the JSON.

The JSON object MUST contain three keys: overall_score, and metrics.
- overall_score: the simple average of all the metric scores
- metrics: a dictionary containing the required metrics and their scores (Type: Object)

Required JSON Schema:

{
    "overall_score": 0.0,
    "metrics": {
    "subject_consistency": 0,
    "aesthetic_quality": 0,
    "temporal_flickering": 0,
    },
}

Your response to the following video input MSUT be this JSON object and nothing else.

\end{Verbatim}

\section{Visual Results}
\label{sup:visual}

\cref{fig:fig:sup_wan_i_to_v,fig:fig:sup_wan_t_to_v,fig:fig:sup_hunyuan_t_to_v} show additional qualitative comparisons on Wan2.2 Image-to-Video, Wan2.2 Text-to-Video, and HunyuanVideo Text-to-Video. Across all examples, Full Attention, Ours, and Ours-Flash produce nearly indistinguishable generations, preserving the same scene layout, subject appearance, and motion patterns. These results show that our sparse attention variants maintain the visual fidelity of full attention while enabling more efficient inference.

\figSupWanItoV
\figSupWanTtoV
\figSupHunyuanTtoV

\end{document}